\documentclass[letterpaper]{article} 
\usepackage{aaai2026}  
\usepackage{times}  
\usepackage{helvet}  
\usepackage{courier}  
\usepackage[hyphens]{url}  
\usepackage{graphicx} 
\usepackage{natbib}  
\usepackage{caption} 
\usepackage{algorithm}
\usepackage{algorithmic}
\usepackage{pifont}
\usepackage{enumitem}
\usepackage{amssymb}
\usepackage{amsmath}
\usepackage{makecell}

\usepackage{array}          
\usepackage{booktabs}       
\usepackage{multirow}       
\usepackage{caption}        
\usepackage[table]{xcolor}  
\usepackage{graphicx} 
\usepackage{subfigure} 

\usepackage{subcaption}  
\usepackage{colortbl}    
\usepackage{mathtools}

\usepackage{newfloat}
\usepackage{listings}
\DeclareCaptionStyle{ruled}{labelfont=normalfont,labelsep=colon,strut=off} 
\floatstyle{ruled}
\newfloat{listing}{tb}{lst}{}
\floatname{listing}{Listing}

\title{\centering\fontsize{14pt}{17pt}\selectfont\bfseries FineTec: Fine-Grained Action Recognition Under Temporal Corruption via Skeleton Decomposition and Sequence Completion}
\author {
    Dian Shao\textsuperscript{\rm 1}\thanks{Corresponding author.},
    Mingfei Shi\textsuperscript{\rm 1},
    Like Liu\textsuperscript{\rm 2}
}
\affiliations {
    \textsuperscript{\rm 1}Unmanned System Research Institute, Northwestern Polytechnical University, Xi'an, China\\
    \textsuperscript{\rm 2}School of Software, Northwestern Polytechnical University, Xi'an, China\\
    shaodian@nwpu.edu.cn, \{mingfeishi5, like.liu\}@mail.nwpu.edu.cn
}

\begin{document}

\maketitle

\begin{abstract}
Recognizing fine-grained actions from temporally corrupted skeleton sequences remains a significant challenge, particularly in real-world scenarios where online pose estimation often yields substantial missing data.
Existing methods often struggle to accurately recover temporal dynamics and fine-grained spatial structures, resulting in the loss of subtle motion cues crucial for distinguishing similar actions.
To address this, we propose \textbf{FineTec}, a unified framework for \textbf{Fine}-grained action recognition under \textbf{Te}mporal \textbf{C}orruption. 
FineTec first restores a base skeleton sequence from corrupted input using context-aware completion with diverse temporal masking. 
Next, a skeleton-based spatial decomposition module partitions the skeleton into five semantic regions, further divides them into dynamic and static subgroups based on motion variance, and generates two augmented skeleton sequences via targeted perturbation.
These, along with the base sequence, are then processed by a physics-driven estimation module, which utilizes Lagrangian dynamics to estimate joint accelerations.
Finally, both the fused skeleton position sequence and the fused acceleration sequence are jointly fed into a GCN-based action recognition head.
Extensive experiments on both coarse-grained (NTU-60, NTU-120) and fine-grained (Gym99, Gym288) benchmarks show that FineTec significantly outperforms previous methods under various levels of temporal corruption. Specifically, FineTec achieves top-1 accuracies of 89.1\% and 78.1\% on the challenging Gym99-severe and Gym288-severe settings, respectively, demonstrating its robustness and generalizability.
Code and datasets could be found at {https://smartdianlab.github.io/projects-FineTec/}.

\end{abstract}


\section{Introduction}

\begin{figure}[t]
    \centering
    \includegraphics[width=1.0\linewidth]{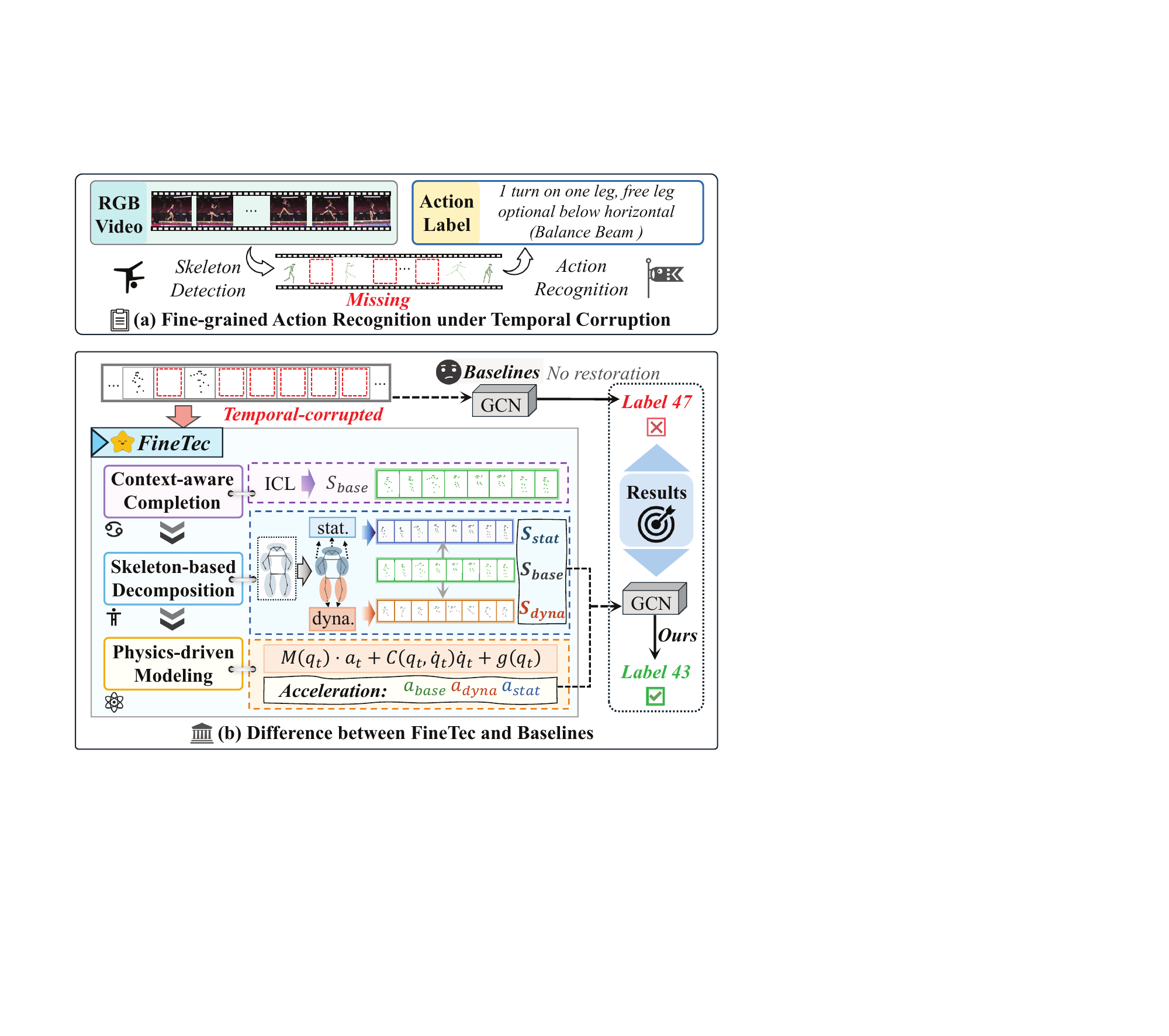}
    \caption{(a) Illustration of the challenging task: Fine-Grained Action Recognition under Temporal Corruption. 
    (b) Compared to other GCN-based methods, the proposed FineTec framework can restore corrupted skeleton sequences and extract more discriminative features for recognition through context-aware completion, skeleton-based decomposition, and physics-driven modeling.
    }
    \label{fig:fig1_motivation}
\end{figure}

Fine-grained action recognition (FAR) aims to identify human actions characterized by slight temporal variations and subtle semantic differences, making it a particularly challenging problem~\cite{shao2020FineGym}.
Skeleton-based representations have emerged as an effective modality for FAR due to their compactness and explicit focus on motion cues~\cite{duan2022revisiting(PoseC3D)}.
However, in complex scenarios such as gymnastics, e.g., \textit{``salto forward stretched with 2 twists''}~\cite{shao2020FineGym}, online pose estimation can suffer from severe frame dropping, reaching up to 69.6\% dropping rate during rapid motion~\cite{zheng2024nettrack}. This results in temporally corrupted skeletal sequences and substantial performance degradation, as shown in Figure~\ref{fig:fig1_motivation}. The problem is especially critical for FAR, which depends on subtle, continuous motion cues~\cite{huang2025sefar, myung2024degcn} and is thus highly sensitive to temporal discontinuities.

Nevertheless, current skeleton-based approaches mainly face two limitations when handling FAR under temporal corruption: 
(1) inadequate temporal recovery, as most models are trained on clean, offline-annotated skeleton sequences and lack mechanisms to handle online detection artifacts~\cite{liu2025revealing, jiang2024lighter, xie2024dynamic}; and
(2) insufficient spatiotemporal modeling, as they often overlook the inherent biological structure of the human body, focusing primarily on point-wise positional features while neglecting continuous kinematic constraints~\cite{li2022dynamic, leong2022combined, chi2022infogcn}.

To address these challenges, we introduce \textbf{FineTec}, a unified framework for \textbf{Fine}-grained action recognition under \textbf{Te}mporal \textbf{C}orruption, as shown in Figure~\ref{fig:2_overview}.
Specifically, FineTec consists of three key modules:
(1) The \textit{Context-aware Sequence Completion} module restores severely corrupted skeleton sequences via diverse temporal masking and in-context learning strategies, enabling an approximate recovery of missing frames and temporal continuity.
(2) The \textit{Skeleton-based Spatial Decomposition} module partitions skeleton joints into five semantic regions based on biological priors, and further divides them into dynamic and static subgroups by motion variance. Targeted augmentation techniques are then applied within each subgroup to generate two sequences, amplifying fine-grained action distinctions.
(3) The \textit{Physics-driven Acceleration Modeling} module re-estimates joint accelerations at each time step using Lagrangian dynamics and pseudo-acceleration (computed as temporal differences between frames). The resulting acceleration sequences effectively capture discriminative motion cues essential for FAR.
The whole process of how the skeleton sequence is restored, processed and utilized is illustrated in Figure~\ref{fig:fig1_motivation}.
Finally, the fused skeleton sequence and its corresponding acceleration cues are integrated for action recognition via a GCN-based network.

To enable comprehensive evaluation, we construct the Gym288-skeleton dataset by extending the open-source Gym99. We manually annotate 11,000 initial-frame bounding boxes, apply OSTrack for athlete tracking, and perform pose estimation within the tracked boxes. The resulting dataset provides skeleton annotations for 288 fine-grained action classes, offering a more challenging benchmark for future research.
To validate the effectiveness of FineTec, we conduct comprehensive experiments on both coarse-grained (NTU-60, NTU-120) and fine-grained (Gym99-skeleton, Gym288-skeleton) datasets, systematically simulating varying levels of temporal corruption, including minor (25\% frame drop), moderate (50\%), and severe (75\%). Experimental results demonstrate that FineTec consistently outperforms previous methods across all settings, and maintains strong recognition accuracy even under severe frame-dropping scenarios.

Our contributions are summarized as follows:
\begin{itemize}[leftmargin=*]
\item We formalize and benchmark fine-grained action recognition under temporal corruption, constructing a large-scale dataset, \textit{Gym288-skeleton}, for comprehensive evaluation;
\item We propose \textbf{FineTec}, a unified framework that integrates context-aware sequence completion, biologically-aware spatial decomposition, and physics-driven temporal refinement to recover corrupted temporal continuity and enhance skeleton sequence quality for improved recognition;
\item Through extensive experiments on both coarse- and fine-grained datasets, we demonstrate that FineTec achieves state-of-the-art performance, especially in the presence of severe temporal corruption.
\end{itemize}

\section{Related Work}

\subsection{Skeleton-based Fine-grained Action Recognition}

Fine-grained Action Recognition (FAR) aims to distinguish subtle action differences (\textit{e.g.}, ``pike sole circle backward with 0.5 turn to handstand''), enabling specialized analysis beyond coarse-grained categories~\cite{zhang2021temporal,yang2020temporal, wang2018temporal,FineQuest,shao2020intra,rajendran2024review}. 
Among modalities, skeletons provide an effective FAR representation by capturing human dynamics while avoiding background noise. 
And the key to FAR is enhancing distinctions in subtle motion details: 
MDR-GCN~\cite{liu2023multidimensional} and Sparse~\cite{xie2025spatial} enhance skeletal features multi-dimensionally.
BlockGCN~\cite{zhou2024blockgcn} optimizes topological structure of GCNs to obtain more discriminative features.
PoseConv3D~\cite{duan2022revisiting(PoseC3D)} employs a heatmap for spatio-temporal dynamics. 
PGVT~\cite{zhang2024pgvt} and SCoPLe~\cite{zhu2025semantic} integrates keypoint with multi-modal features.
However, they predominantly emphasize ``displacement'' information, relying on data-driven methods to implicitly learn complex temporal dynamics, and lacking guidance from physical realism.
While ActCLR~\cite{lin2023actionlet} targets fine-grained resolution in skeletal space, its contrastive learning framework limits comprehensive exploitation.
Diverging from prior work, FineTec enables physically interpretable modeling of fine-grained actions by concurrently addressing skeletal spatial granularity and physics-constrained temporal dynamics.

\begin{figure*}[!t]
    \centering
    \includegraphics[width=1.0\linewidth]{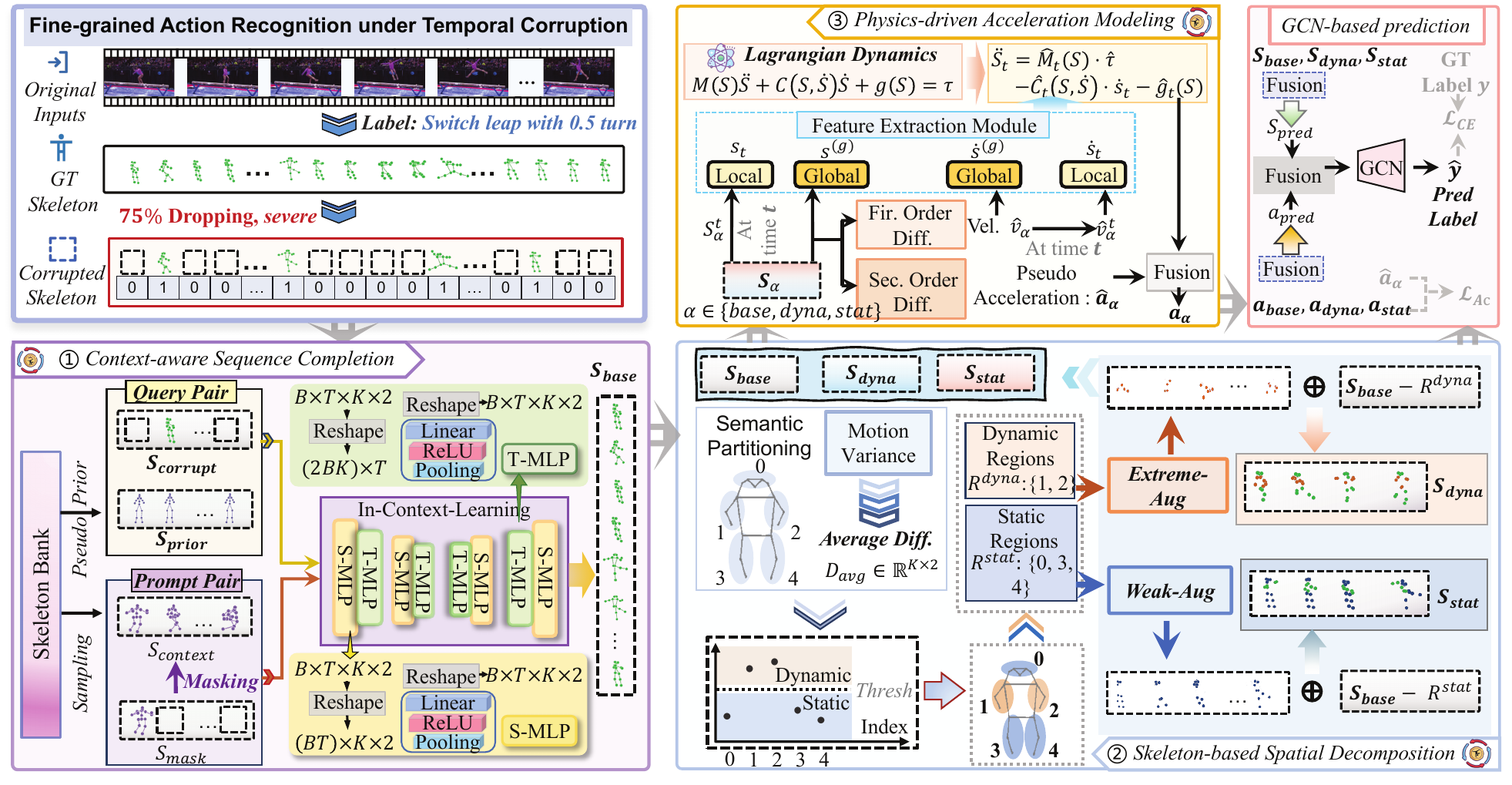}
    \caption{Overview of the Pipeline.
    FineTec consists of three core modules:
    \ding{172} Context-aware Sequence Completion restores missing or corrupted skeleton frames using in-context learning, producing $S_{base}$; \ding{173} Skeleton-based Spatial Decomposition partitions $S_{base}$ into anatomical regions by motion intensity, generating dynamic ($S_{dyna}$) and static ($S_{stat}$) variants, which are fused into $S_{pred}$; \ding{174} Physics-driven Acceleration Modeling infers joint accelerations via Lagrangian dynamics and data-driven finite differences, producing fused temporal dynamics features $\mathbf{a}$. The resulting positional ($S_{pred}$) and dynamic (${a}_{pred}$) features are used for downstream fine-grained action recognition.
    }
    \label{fig:2_overview}
\end{figure*}

\subsection{Physics-aware Video Understanding}

Traditional video understanding is fundamentally limited by the agnostic nature of models and weakly interpretable feature spaces~\cite{lin2025exploring, wang2025action,Shao_2025_CVPR}. 
And physics-aware manners~\cite{gartner2022differentiable, gartner2022trajectory} offers a promising alternative by embedding physical principles. 
Some works leverage simulation environments for phenomena like rigid body collision or fluid dynamics~\cite{andriluka2025learned, liu2024physgen}, but adjusting physical parameters within engines remains challenging. 
Others directly integrate mathematical physics equations into model design~\cite{zhang2024physpt, ugrinovic2024multiphys}.
Among them, PIMNet~\cite{zhang2022pimnet} uses Newtonian for motion prediction, InfoGCN++~\cite{chi2024InfoGCN++} utilizes Neural-ODEs for action recognition, and LieGroupHamDL~\cite{duong23porthamiltonian} combines Lie groups with Hamiltonian for robot control. 
In contrast, our FineTec: 
(1) focuses on fine-grained temporal-corrupted action recognition tasks; 
(2) introduces the ICL mechanism and masking strategies for temporal completion;
(3) and integrates biological priors and Lagrangians for fine-grained analysis.

\section{Methodology}
\subsection{Preliminary}
\subsubsection{\ding{111} Physics of Rigid-body Dynamics.} 
Rigid-body dynamics are typically formulated using Lagrangian or Hamiltonian methods, with the Lagrangian formulation~\cite{zhang2024incorporating} widely adopted for its clarity and simplicity. 
For human kinematics~\cite{jain1995diagonalized}, the Lagrangian dynamics are described by:
\begin{equation} \label{eq:lagrangian}
    M(q_t)\cdot \ddot{q}_t + C(q_t, \dot{q}_t)\dot{q}_t + g(q_t) = \tau_t,
\end{equation}
where $q_t$ and $\dot{q}_t$ denote generalized coordinates, at time $t$.
$M(q_t)$ is the configuration-dependent inertia matrix reflecting the mass distribution of body segments; $C(q_t, \dot{q}_t)\dot{q}_t$ represents Coriolis and centrifugal forces; $g(q_t)$ accounts for gravitational forces; and $\tau_t$ denotes the vector of generalized forces, including joint torques and external influences.

\subsubsection{\ding{109} Task Definition}
In this work, we address the challenging task of fine-grained action recognition under temporal corruption. 
Given a ground-truth 2D skeleton sequence  $S_{gt}$$ \in \mathbb{R}^{T \times K \times 2}$, where $T$ is the number of frames and $K=17$ is the number of joints, we simulate temporal corruption by randomly dropping 25\% (minor), 50\% (moderate), or 75\% (severe) of the frames to obtain $ S_{corrupt}  \in \mathbb{R}^{T \times K \times D}$, with only $\hat{T} < T$ valid frames and the rest zero-padded.
This setting reflects the missing data challenges frequently encountered in real-world online action detection scenarios.
The goal is to predict the action category $y \in \{1, \dots, C\}$ from the corrupted sequence $ S_{corrupt} $, where $C$ is the number of action classes.

\subsection{The FineTec Framework}
\subsubsection{\ding{109} Overall Pipeline.}
The FineTec framework processes temporally corrupted skeleton sequences through three key modules to achieve better fine-grained action recognition results, as illustrated in Figure~\ref{fig:2_overview}.
First, the \textit{Context-aware Sequence Completion} module employs in-context learning to restore missing or corrupted frames, producing a basically completed skeleton sequence $S_{base}$. 
Next, the \textit{Skeleton-based Spatial Decomposition} module partitions $S_{base}$ into five anatomical regions and classifies them by motion intensity. Region-specific augmentations generate dynamic ($S_{dyna}$) and static ($S_{stat}$) variants, which are fused to yield the final sequence $S_{pred}$. 
These processed sequences are then used to extract displacement and acceleration features, which serve as inputs for downstream recognition networks.
Finally, the \textit{Physics-driven Acceleration Modeling} module infers joint accelerations via physics-based Lagrangian dynamics, 
and combine with data-driven finite differences to generate the fused temporal dynamics features $\mathbf{a}$. 
The resulting positional $S_{pred}$ and dynamic features $\mathbf{a}$ are then utilized together for downstream fine-grained action recognition.

\subsubsection{\ding{168} Context-aware Sequence Completion.}
To handle temporally corrupted skeleton sequences $S_{corrupt}$, we adopt an In-Context Learning (ICL) paradigm~\cite{wang2024skeleton, kim2025videoicl} to approximately recover the complete sequence at first. We first construct a skeleton bank from Human3.6M 2D skeleton data, and obtain an average prior sequence $S_{prior}$ via temporal averaging.
For each training instance, a sequence is sampled from the bank and corrupted by one of five temporal masking strategies: random, pattern-based, or contiguous block masking (prefix, suffix, in-between). 
The original $S_{context}$ and the masked sequence $S_{mask}$  form a prompt pair, demonstrating recovery from corruption. 
The input $S_{corrupt}$ is paired with $S_{prior}$ as a query pair.
Both the prompt and query pairs are processed by lightweight spatial and temporal MLPs, enabling the network to approximately restore the base sequence $S_{base}$ by contextually completing missing frames.
Further implementation details are provided in the Appendix.

\subsubsection{\ding{169} Skeleton-based Spatial Decomposition.}
This module enhances fine-grained action discrimination by decomposing and augmenting the predicted skeleton sequence $S_{base}$ based on motion analysis and anatomical priors.
Specifically, leveraging the human biological structure, we first partition the $K$ joints into five semantic regions: head ($G_0$), left arm ($G_1$), right arm ($G_2$), left leg ($G_3$), and right leg ($G_4$). 
To quantify the motion level of each joint, we compute the average frame-wise displacement as follows:
\begin{align}
    D_{avg}^{(i)} = \frac{1}{T-1} \sum_{t=0}^{T-2} \| S_{base}^{t+1, i} - S_{base}^{t, i} \|_2
\end{align} 
where $i$ indexes the joint. The regional motion intensity for each group $G_j$ is then calculated as:
\begin{equation}
    \bar{D}_j = \frac{1}{|G_j|} \sum_{i \in G_j}  D_{avg}^{(i)} \in \mathbb{R}, \quad j \in \{G_1, G_2, \cdots, G_5\}.
\end{equation}

The top two regions with the highest $\bar{D}_j$ are designated as dynamic, and the remaining three as static.
To introduce region-specific diversity, we apply strong spatial-temporal perturbations (such as temporal cropping, random dropping, or interpolation) to $S_{base}$ and selectively replace the \textit{dynamic regions} with their perturbed versions, resulting in $S_{dyna} $. For static regions, we apply only weak spatial perturbations (e.g., random flipping) and substitute the corresponding \textit{static regions} in $S_{base}$, producing $S_{stat}$. 

Finally, $S_{base}$, $S_{dyna}$, and $S_{stat}$ are fused to form the final sequence $S_{pred}$. 
This decomposition-augmentation-fusion strategy preserves temporal coherence, amplifies motion cues in dynamic regions, and stabilizes static postures, collectively improving fine-grained action recognition.

\begingroup
\setlength{\tabcolsep}{1mm}
\small
\begin{table*}[t]
  \centering
  \begin{tabular}{lccccccccccccc}
    \toprule
    & & \multicolumn{2}{c}{\textbf{G288-Min.}} & \multicolumn{2}{c}{\textbf{G288-Mod.}} & \multicolumn{2}{c}{\textbf{G288-Sev.}} & \multicolumn{2}{c}{\textbf{G99-Min.}} & \multicolumn{2}{c}{\textbf{G99-Mod.}} & \multicolumn{2}{c}{\textbf{G99-Sev.}}  \\
    \cmidrule(lr){3-4}\cmidrule(lr){5-6}\cmidrule(lr){7-8}\cmidrule(lr){9-10} \cmidrule(lr){11-12} \cmidrule(lr){13-14}
    \multirow{-3}{*}{\textbf{Method}} & \multirow{-3}{*}{\textbf{Input}} & \textbf{Top-1} & \textbf{Mean} & \textbf{Top-1} & \textbf{Mean} & \textbf{Top-1} & \textbf{Mean} & \textbf{Top-1} & \textbf{Mean} & \textbf{Top-1} & \textbf{Mean} & \textbf{Top-1} & \textbf{Mean} \\
    \midrule
    ST-GCN $\textit{\textcolor[RGB]{130,130,130}{AAAI'18}}$ & Skeleton & 0.784 & 0.381 & 0.770 & 0.344 & 0.742 & 0.304 & 0.895 & 0.869 & 0.876 & 0.829 & 0.871 & 0.783  \\
    PYSKL-J $\textit{\textcolor[RGB]{130,130,130}{arXiv'22}}$ & Skeleton & 0.813 & 0.401 & 0.794 & 0.368 & 0.773 & 0.315 & 0.920 & 0.871 & 0.903 & 0.856 & 0.884 & 0.791  \\
    PYSKL-B $\textit{\textcolor[RGB]{130,130,130}{arXiv'22}}$ & Skeleton & 0.811 & 0.385 & 0.796 & 0.373 & 0.765 & 0.314 & 0.915 & 0.872 & 0.905 & 0.858 & 0.889 & 0.801  \\
    PoseC3D-J $\textit{\textcolor[RGB]{130,130,130}{CVPR'22}}$ & Heatmap & 0.793 & 0.297 & 0.771 & 0.284 & 0.747 & 0.253 & 0.916 & 0.871 & 0.904 & 0.854 & 0.873 & 0.770  \\
    PoseC3D-L $\textit{\textcolor[RGB]{130,130,130}{CVPR'22}}$ & Heatmap & 0.790 & 0.296 & 0.775 & 0.281 & 0.756 & 0.250 & 0.917 & 0.870 & 0.899 & 0.848 & 0.870 & 0.761  \\
    AAGCN $\textit{\textcolor[RGB]{130,130,130}{TIP'20}}$ & Skeleton & 0.755 & 0.281 & 0.765 & 0.279 & 0.744 & 0.263 & 0.907 & 0.856 & 0.902 & 0.846 & 0.874 & 0.795  \\
    CTRGCN $\textit{\textcolor[RGB]{130,130,130}{ICCV'21}}$ & Skeleton & 0.786 & 0.292 & 0.784 & 0.285 & 0.760 & 0.271 & 0.914 & 0.874 & 0.897 & \textbf{0.859} & 0.884 & 0.803  \\
    Sparse $\textit{\textcolor[RGB]{130,130,130}{CVPR'25}}$ & Skeleton & 0.765 & 0.282 & 0.740 & 0.268 & 0.683 & 0.237 & 0.898 & 0.860 & 0.876 & 0.827 & 0.808 & 0.725  \\
    \midrule
    \rowcolor{gray!20}
    \textbf{FineTec (Ours)} & Skeleton & \textbf{0.815} & \textbf{0.404}  & \textbf{0.797} & \textbf{0.381} & \textbf{0.781} & \textbf{0.356} & \textbf{0.921} & \textbf{0.875} & \textbf{0.906} & 0.851 & \textbf{0.891} & \textbf{0.805}  \\
    \bottomrule
  \end{tabular}
\caption{Fine-grained action recognition on Gym99-skeleton and Gym288-skeleton. Both Top-1 accuracy and mean class accuracy are reported under minor (Min.), moderate (Mod.), and severe (Sev.) temporal corruption.}
\label{tab:main_results_fine}
\end{table*}
\endgroup

\subsubsection{\ding{170} Physics-driven Acceleration Modeling.}
In this module, we explicitly model acceleration dynamics using Lagrangian principles to enhance motion representation.

Recall the Lagrangian equation (Eq.~\ref{eq:lagrangian}), where we substitute the general coordinates $q$ with the set of joint positions $S$:
\begin{equation} 
    M(S)\ddot{S} + C(S,\dot{S})\dot{S} + g(S) = \tau,
\end{equation}
where $M$, $C$, $g$, and $\tau$ denote the inertia matrix, Coriolis matrix, gravity term, and driving force, respectively.
Here $S$ could be $S_{base}$, $S_{dyna}$, and $S_{stat}$, and we omit the subscripts for convenience.
Our aim is to calculate the acceleration term $\ddot{S}$:
\begin{equation}
    \ddot{S} = \{M(S)\}^{-1}\cdot\tau - \hat{C}(S,\dot{S})\dot{S} - \hat{g}(S).
\end{equation}
For computational efficiency and to facilitate neural network estimation, we define: $\hat{C}(S,\dot{S}) \coloneqq {M(S)}^{-1} C(S,\dot{S})$, $\hat{g}(S) \coloneqq {M(S)}^{-1} g(S)$, and $\hat{M}(S) \coloneqq {M(S)}^{-1}$.
To provide the necessary inputs for these estimators, we first extract both global and local features of the joint positions and velocities:
\begin{align} 
    s^{(g)} &= f_{s}^{(global)}(\{S^t\}_{t=0}^{T-1}), \;  s_{t} = f_s^{(local)}(S_t); \\ 
     \dot{s}^{(g)} &= f_{\dot{s}}^{(global)}(\{\hat{v}_t\}_{t=0}^{T-1}),  \; \dot{s_t} = f_{\dot{s}}^{(local)}(\hat{v}_t);
\end{align} 
where $\hat{v}_{\alpha}^{t}$ is calculated using first-order finite differences as
$\hat{v}_{t} = \frac{S_{t+1} - S_{t-1}}{2\Delta t}$.
Each physical term in the dynamics equation is then estimated using neural networks $\mathbb{E}$. At a specific time t:
\begin{align}
  \hat{g_t}(S) &= \mathbb{E}_{g} [s^{(g)}, s_t]. 
\end{align}
 Since $\tau$ is time-independent, we have:
 \begin{equation}
     \hat{\tau} = \mathbb{E}_{\tau} [s^{(g)}, \dot{s}^{(g)}]
 \end{equation}
The remaining two terms are matrices, which we assume to be symmetric. Therefore, we first estimate their upper triangular parts and obtain the final matrices by applying the symmetry operation $\mathcal{S}^{\text{\dag}}$:
\begin{align}
     \hat{C_t}(S,\dot{S}) &= \mathcal{S}^{\text{\dag}} \{ \mathbb{E}_{C} [s^{(g)}, s_t, \dot{s}^{(g)}, \dot{s_t}] \}, \\
   \hat{M_t}(S) &= \mathcal{S}^{\text{\dag}} \{ \mathbb{E}_{M} [s^{(g)}, s_t] \}. 
\end{align}
The refined, physics-driven acceleration is then computed as:
\begin{align} 
    \ddot{S}_t &= \hat{M}_t(S) \cdot \hat{\tau} - \hat{C}_{t}(S,\dot{S}) \cdot \dot{s_t} - \hat{g}_t(S).
\end{align}
To further improve robustness, we combine this estimate with the pseudo-acceleration calculated via second-order finite differences:
$\hat{a}_t = \frac{S_{t+1} - 2S_{t} + S_{t-1}}{(\Delta t)^2}$.
The final fused temporal dynamics feature (joint acceleration) is:
\begin{equation}
     \textbf{a}_t = \text{Fusion}(\hat{a}_t, \ddot{S}_t) \in \mathbb{R}^{K \times 2}.
\end{equation}

\begingroup
\setlength{\tabcolsep}{1mm}
\footnotesize
\begin{table}[t]
  \centering
  \begin{tabular}{lcccccc}
    \toprule
    \textbf{Method} & \multicolumn{3}{c}{\textbf{NTU-60}} & \multicolumn{3}{c}{\textbf{NTU-120}} \\
    \cmidrule(lr){2-4} \cmidrule(lr){5-7}
    & \textbf{Min.} & \textbf{Mod.} & \textbf{Sev.} & \textbf{Min.} & \textbf{Mod.} & \textbf{Sev.} \\
    \midrule
    ST-GCN & 0.894 & 0.890 & 0.879 & 0.810 & 0.803 & 0.781 \\
    PYSKL-J & 0.885 & 0.883 & 0.875 & 0.809 & 0.808 & 0.790 \\
    PYSKL-B  & 0.893 & 0.887 & 0.885 & 0.815 & 0.810 & 0.790 \\
    PoseC3D-J & 0.887 & 0.889 & 0.878 & \textbf{0.823} & 0.795 & 0.783 \\
    PoseC3D-L & 0.899 & 0.897 & 0.877 & 0.816 & 0.812 & 0.785 \\
    AAGCN & 0.891 & 0.886 & 0.873 & 0.813 & 0.807 & 0.796 \\
    CTRGCN & 0.901 & 0.892 & 0.879 & 0.814 & 0.809 & 0.793 \\
    Sparse & 0.895 & 0.896 & 0.864 & 0.813 & 0.793 & 0.767 \\
    \midrule
    \rowcolor{gray!20}
    \textbf{FineTec (Ours)} & \textbf{0.903} & \textbf{0.901} & \textbf{0.892} & 0.819 & \textbf{0.817} & \textbf{0.813} \\
    \bottomrule
  \end{tabular}
\caption{Coarse-grained action recognition on NTU-60-xsub and NTU-120-xsub. Top-1 accuracy is reported under minor, moderate, and severe temporal corruption.}
\label{tab:main_results_coarse}
\end{table}
\endgroup

\subsubsection{\ding{171} GCN-based Optimization Objectives.}
The FineTec framework is trained in two stages: skeleton sequence completion and the action recognition task.

\noindent\ding{192} For sequence completion, we use mean squared error (MSE) losses to measure the difference between the completed skeleton and the ground-truth sequence, for both the prompt and query pairs:
\begin{equation}
    \mathcal{L}_{ICL}=\text{MSE}(S_{gt},S_{base}) + \text{MSE}(S_{context},S_{mask})
\end{equation}

\noindent\ding{193} For action recognition, the model utilizes the fused skeleton sequence $S_{pred}$ and fused acceleration sequence $\mathbf{a}_{pred}$, which are integrated through a cross-attention module to capture both positional and dynamic information. The resulting representations are processed by graph convolutional networks (GCNs)~\cite{duan2022pyskl}, followed by a recognition head that predicts class probabilities $\hat{y}$.
The classification loss is defined as the cross-entropy between the predicted and ground-truth labels:
\begin{equation}
\mathcal{L}_{CE} = -\sum_{i} y_i \log \hat{y}_i,
\end{equation}
where $y_i$ is the ground-truth label.
An additional loss is also calculated between $\hat{\mathbf{a}}$ and $\mathbf{a}$ over the three sequences:
\begin{equation}
\mathcal{L}_{Ac} = \frac{1}{3} \sum_{\alpha} \text{MSE}(\hat{\mathbf{a}}_{\alpha}, \mathbf{a}_{\alpha}), \quad \alpha \in \{base, dyna, stat \}.
\end{equation}
The overall training objective is:
\begin{equation}
\mathcal{L} = \mathcal{L}_{CE} + \lambda \mathcal{L}_{Ac},
\end{equation}
where $\lambda$ is a balancing hyperparameter.

\begin{figure}[t]
    \centering
    \includegraphics[width=1.0\linewidth]{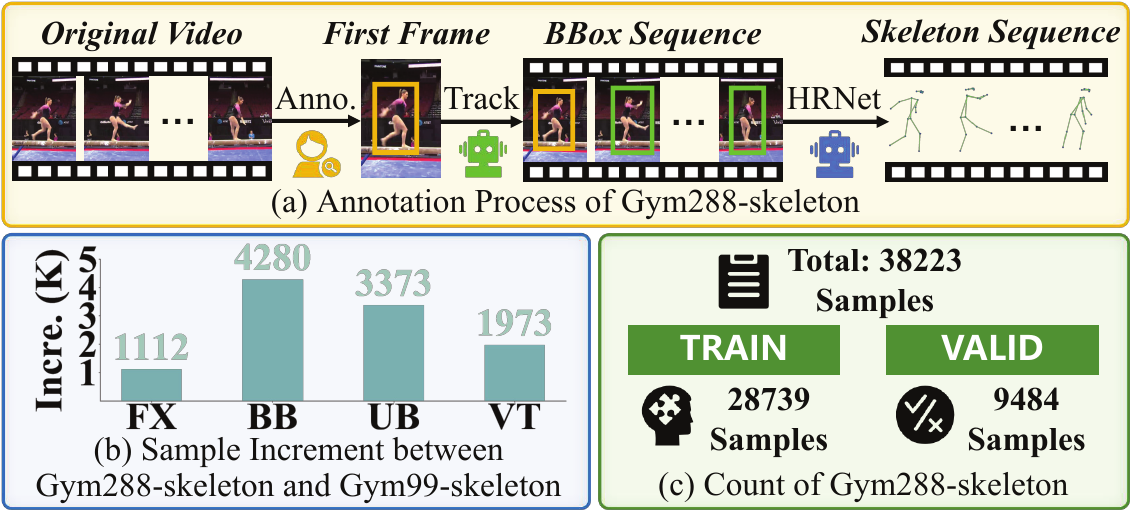}
    \caption{The Construction Process and Statistics of the constrcucted Gym288-skeleton Dataset.}
    \label{fig:gym288_skeleton_statistics}
\end{figure}

\begingroup
\setlength{\tabcolsep}{1mm}
\begin{table*}[ht]
  \centering
  \begin{tabular}{lccccccccc}
    \toprule
    & \multicolumn{3}{c}{\textbf{Gym99-Min.}} & \multicolumn{3}{c}{\textbf{Gym99-Mod.}} & \multicolumn{3}{c}{\textbf{Gym99-Sev.}} \\
    \cmidrule(lr){2-4}\cmidrule(lr){5-7}\cmidrule(lr){8-10}
    \multirow{-2}{*}{\textbf{Method}}  
    & \textbf{MPJPE$\downarrow$} & \textbf{N-MPJPE$\downarrow$} & \textbf{MPJVE$\downarrow$}
    & \textbf{MPJPE$\downarrow$} & \textbf{N-MPJPE$\downarrow$} & \textbf{MPJVE$\downarrow$}
    & \textbf{MPJPE$\downarrow$} & \textbf{N-MPJPE$\downarrow$} & \textbf{MPJVE$\downarrow$} \\
    \midrule
    L-R Copy 
    & 0.136 & 0.133 & 0.246
    & 0.332 & 0.318 & 0.445
    & 0.713 & 0.665 & 0.575 \\
    R-L Copy 
    & 0.136 & 0.132 & 0.246
    & 0.327 & 0.313 & 0.441
    & 0.699 & 0.650 & 0.571 \\
    \midrule
    siMLPe 
    & 0.175 & 0.139 & 0.119
    & 0.208 & 0.168 & 0.129
    & 0.245 & 0.199 & 0.139 \\
    SiC-Stat 
    & 0.210 & 0.102 & 0.351
    & 0.397 & 0.181 & 0.508
    & 0.584 & 0.252 & 0.544 \\
    SiC-Dyna
    & 0.188 & 0.100 & 0.196
    & 0.164 & 0.144 & 0.205
    & 0.192 & 0.174 & 0.321 \\
    \midrule
    \rowcolor{gray!20}
    \textbf{Ours} 
    & \textbf{0.106} & \textbf{0.098} & \textbf{0.047}
    & \textbf{0.119} & \textbf{0.109} & \textbf{0.085}
    & \textbf{0.147} & \textbf{0.132} & \textbf{0.113} \\
    \bottomrule
  \end{tabular}
\caption{Skeleton restoration results on Gym99-Skeleton.
Experiments are conducted under three levels of temporal corruption: minor, moderate, and severe. The evaluation metrics include MPJPE, N-MPJPE, and MPJVE. And the ``L-R'' and ``R-L'' denote left-to-right and right-to-left respectively.}
\label{tab:main_results_pose_completion}
\end{table*}
\endgroup

\section{Experiment}

\begin{table}[t]
\centering
\setlength{\tabcolsep}{1.5mm}
{\small
\begin{tabular}{c c c c c c c}
  \hline
  \textbf{Index} & \textbf{Cont.} & \textbf{Skel.} & \textbf{Phys.} & \textbf{Minor} & \textbf{Moderate} & \textbf{Severe} \\
  \hline
  \ding{182} & \ding{55} & \ding{51} & \ding{51} & 0.812 & 0.785 & 0.751 \\
  \ding{183} & \ding{51} & \ding{55} & \ding{51} & 0.787 & 0.780 & 0.770 \\
  \ding{184} & \ding{51} & \ding{51} & \ding{55} & 0.789 & 0.776 & 0.775 \\
  \hline
  \rowcolor{gray!20}
  \ding{185} & \ding{51} & \ding{51} & \ding{51} & \textbf{0.815} & \textbf{0.797} & \textbf{0.781} \\
  \hline
\end{tabular}
}
\caption{ Ablations of different modules, including the Context-aware Sequence Completion  (Cont.), Skeleton-based Spatial Decomposition (Skel.), and Physics-driven Acceleration Modeling (Phys.). }
\label{tab:ablations}
\end{table}
\begin{table}[t]
\centering
\setlength{\tabcolsep}{3.5mm}
{\small
\begin{tabular}{c c c| c c}
  \hline
  \textbf{Index} & \textbf{$S_{dyna}$} & \textbf{$S_{stat}$} & \textbf{Moderate} & \textbf{Severe} \\
  \hline
  \ding{182} & \ding{55} & \ding{51} & 0.790 & 0.774 \\
  \ding{183} & \ding{51} & \ding{55} & 0.786 & 0.764 \\
  \hline
  \rowcolor{gray!20}
  \ding{184} & \ding{51} & \ding{51} & \textbf{0.797} & \textbf{0.781} \\
  \hline
\end{tabular}
}
\caption{Analysis on augmented skeleton sequences.}
\label{tab:ablation_skeleton}
\end{table}

\subsection{Gym288-Skeleton Dataset}
In this work, to enable comprehensive evaluation of fine-
grained and temporally corrupted action recognition, we
construct the Gym288-skeleton dataset by extending the
open-source Gym99 dataset. We manually annotate $\sim 1.1$w
initial-frame bounding boxes, apply OSTrack (Ye et al.
2022) for athlete tracking, and perform pose estimation
within the tracked boxes for each video. This provides
a large-scale dataset with skeleton annotations for 288
fine-grained action classes, advancing the scope and difficulty of existing benchmarks. 
Statistics are shown in Figure~\ref{fig:gym288_skeleton_statistics}, and the detailed constructing process and analysis are shown in the supplementary material.

\subsection{Evaluation Settings}
\ding{245} \textbf{Datasets:} We conduct fine-grained action recognition experiments on two benchmark datasets: Gym99-skeleton and the constructed Gym288-skeleton~\cite{shao2020FineGym}. For coarse settings, we use the NTU datasets, including NTU60~\cite{shahroudy2016ntu} and NTU120~\cite{liu2020ntu}.
\ding{245} \textbf{Baselines:} We compare our method with several representative skeleton-based approaches: ST-GCN~\cite{yan2018ST-GCN}, PYSKL~\cite{duan2022pyskl}, PoseC3D~\cite{duan2022revisiting(PoseC3D)}, AAGCN~\cite{shi2020skeleton}, CTRGCN~\cite{chen2021channel}, and Sparse~\cite{xie2025spatial}.
\ding{245} \textbf{Evaluation Metrics:} For action recognition, we report Top-1 and Top-5 accuracy. Due to the significant class imbalance in Gym288~\cite{shao2020FineGym}, Mean class accuracy (Mean) is also included as a more informative metric. For skeleton restoration, we employ standard metrics: Mean Per Joint Position Error (MPJPE) and Mean Per Vertex Position Error (MPVPE).
Main results are presented in this section, with further details provided in the Appendix.

\subsection{Main Results}

\noindent \textbf{Results on Fine-grained Action Recognition.}
The main quantitative results on the two fine-grained skeleton datasets, Gym99 and Gym288-skeleton, are presented in Table~\ref{tab:main_results_fine}.
These results are reported across three difficulty levels: minor (25\% frame missing), moderate (50\% frame missing), and severe (75\% frame missing). 
It can be observed that the proposed FineTec framework consistently achieves the best performance under all conditions. 
Notably, in the most challenging scenario—Gym288-skeleton with severe frame missing—FineTec attains a Top-1 accuracy of 78.1\%, surpassing all previous skeleton-based methods.
In terms of mean class accuracy, FineTec improves upon the best baseline by 13\%, and outperforms the latest work~\cite{xie2025spatial} by 50\%.
Overall, these results demonstrate that FineTec achieves outstanding effectiveness across fine-grained datasets and under all levels of difficulty.

\noindent \textbf{Results on Coarse-grained Action Recognition.} 
To further evaluate the generalization and robustness of FineTec, we conduct experiments on the coarse-grained skeleton action recognition benchmarks, including NTU-60-xsub and NTU-120-xsub, and UCF101~\cite{soomro2012UCF101}. 
The results are shown in Table~\ref{tab:main_results_coarse}, showing that FineTec outperforms all competitive baselines, particularly under the most challenging (severe) condition, achieving Top-1 accuracy improvements of 1.3\% on NTU-60-xsub and 1.7\% on NTU-120-xsub.
Results on UCF101 are provided in the Appendix due to space limitations.

\noindent \textbf{Main Results on Skeleton Restoration.}
We quantitatively evaluate the ability of FineTec to restore temporally corrupted skeleton sequences, as presented in Table~\ref{tab:main_results_pose_completion}.
Notably, our method achieves the lowest MPJPE across all corruption levels, substantially outperforming all baselines.
Compared to the strongest competing method (SiC-Dyna), MaskICL achieves MPJPE reductions of 43.6\% under minor corruption, 27.4\% under moderate corruption, and 23.4\% under severe corruption.
Consistent improvements are also observed for N-MPJPE and MPJVE, where our method achieves the best performance across all settings.

\subsection{Ablations and Analysis}

\subsubsection{Ablation Studies of Modules.}
We validate FineTec’s design through ablation studies of its three main modules and their variants on the Gym288-skeleton dataset, evaluated in terms of top-1 accuracy.
\ding{182} Module Ablation: Table~\ref{tab:ablations} demonstrates that removing any FineTec module results in a clear performance drop, confirming the necessity of each component and the effectiveness of the overall design.
\ding{183}  Analysis of Skeleton Decomposition:
Table~\ref{tab:ablation_skeleton} shows that combining both $S_{dyna}$ and $S_{stat}$ achieves higher accuracy than using either alone, demonstrating that spatial decomposition and differentiated processing enrich skeleton features and enhance fine-grained action recognition.
\ding{184}  Fusion Strategy in Physics-driven module:
We compare cross-attention (CA) fusion and MLP-based integration for integrating acceleration cues. CA consistently outperforms MLP in both moderate (0.797 vs. 0.779) and severe (0.781 vs. 0.771) settings (Top-1 Acc.), validating the effectiveness of the selected fusion strategy.

\begin{figure}[t]
    \centering
    \includegraphics[width=1.0\linewidth]{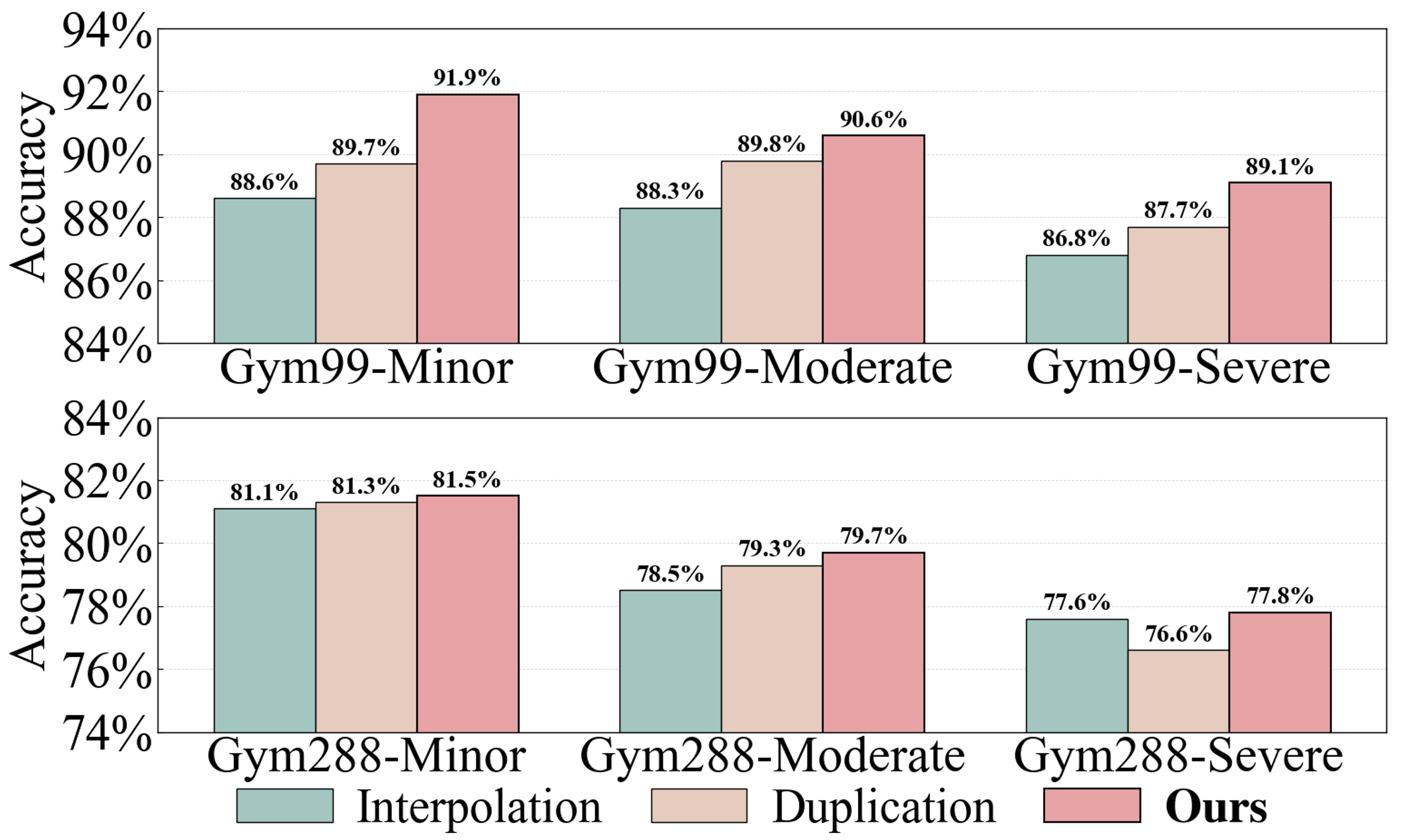}
    \caption{Comparison of Skeleton Restoration Methods on Gym99-skeleton and Gym288-skeleton. Top-1 accuracy of FineTec (Ours), Interpolation, and Duplication is reported under varying levels of temporal corruption.}
    \label{fig:ablation_completion_func}
\end{figure}

\begin{table}[t]
\centering
\setlength{\tabcolsep}{3pt} 
{\small %
    \begin{tabular}{c | c c c c}
      \hline
      \textbf{Perturb.}  & \textbf{G99-Min.} & \textbf{G99-Sev.}  & \textbf{G288-Min.} & \textbf{G288-Sev.} \\
      \hline
      S-low & 0.91/0.985  & 0.842/0.971 & 0.790/0.911 & 0.764/0.905 \\
      S-high & 0.900/0.983 & 0.825/0.967 & 0.785/0.908  & 0.736/0.884 \\
      \hline
      T-low & 0.898/0.988 & 0.869/0.981 & 0.791/0.918  & 0.777/0.915 \\
      T-high & 0.896/0.987 & 0.862/0.979 & 0.789/0.917  & 0.774/0.914 \\
      \hline
      \rowcolor{gray!20}
      \textbf{w/o} & \textbf{0.926/0.995} & \textbf{0.896/0.988} & \textbf{0.819/0.928} & \textbf{0.804/0.924} \\
      \hline
    \end{tabular}%
}
\caption{Robustness analysis under spatial and temporal perturbations. Yellow rows indicate spatial (Gaussian noise) perturbations, and blue rows indicate temporal (frame dropping) perturbations. Top-1 / Top-5 accuracy are reported.}
\label{tab:ablation-robustness-noisy-input}
\end{table}

\subsubsection{Impact of Sequence Completion on Action Recognition.}
We investigate how different sequence completion methods affect fine-grained action recognition accuracy, as shown in Figure~\ref{fig:ablation_completion_func}. Compared to standard approaches such as Interpolation and Duplication, FineTec consistently yields the highest Top-1 accuracy on both fine-grained action recognition datasets, across varying levels of temporal degradation. 
Specifically, on the Gym99 dataset, FineTec achieves Top-1 accuracies of 0.919 (minor), 0.906 (moderate), and 0.885 (severe), demonstrating notable robustness even under significant data loss. The improvements are even more pronounced on the more challenging Gym288 dataset, where FineTec attains Top-1 accuracies of 0.815 (minor), 0.797 (moderate), and 0.778 (severe), outperforming all baselines. In particular, under severe degradation, FineTec significantly surpasses both Duplication and Interpolation.
These results show that effective sequence completion significantly benefits action recognition under temporal corruption.

\subsubsection{Qualitative Visualization of Skeleton Restoration.}
\begin{figure}[t]
    \centering
    \includegraphics[width=1.0\linewidth]{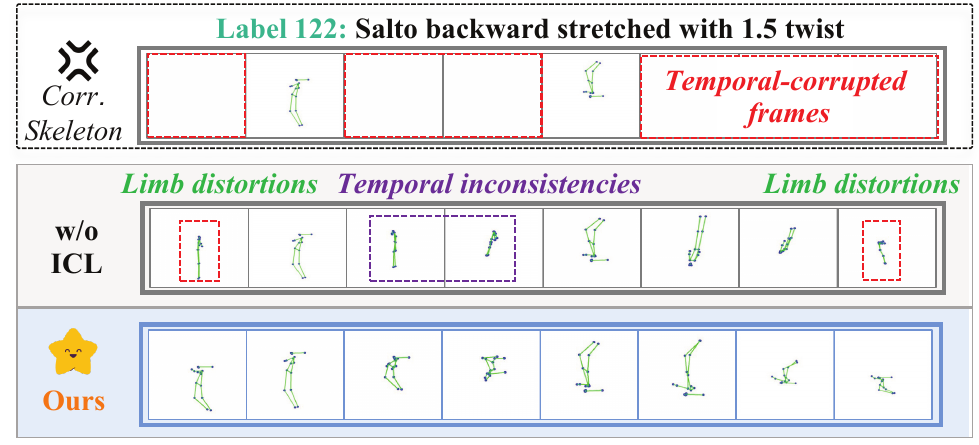}
    \caption{Qualitative Results of Skeleton Restoration. Our context-aware completion method more accurately reconstructs missing frames and preserves fine-grained motion details compared to the ablation without in-context learning.
    }
    \label{fig:ablation_restoring_visualization}
\end{figure}

Figure~\ref{fig:ablation_restoring_visualization} presents a qualitative comparison of skeleton sequence restoration for a sample from the Gym288 dataset (Label 122: \textit{ ``Salto backward stretched with 1.5 twist''}) under severe corruption. We compare our context-aware completion module with an ablation variant that excludes in-context learning. Both methods use identical training settings, yet our approach better reconstructs the missing frames and preserves the fine-grained motion details.

\subsubsection{Robustness to Noisy Inputs.}
We assess the robustness of FineTec under both spatial and temporal perturbations. Spatial noise is introduced by adding Gaussian noise at two severity levels (``S-low'' and ``S-high''). Temporal robustness is evaluated by randomly dropping half of the input frames, also at two severity levels (``T-low'' and ``T-high'').
As summarized in Table~\ref{tab:ablation-robustness-noisy-input}, FineTec maintains high recognition accuracy under all perturbations. 
These results demonstrate FineTec's strong resilience to both spatial and temporal input corruptions, ensuring reliable performance even in challenging real-world scenarios.

\subsubsection{Discussion and Future Works.}

FineTec establishes a strong foundation for robust fine-grained action recognition under temporal corruption, demonstrating the potential of detailed spatio-temporal and physics-driven modeling. Building on this foundation, there remain many promising directions for future research. 
For example, the current use of a fixed skeleton bank in sequence completion and manually defined subgroup partitioning can inspire future work on more adaptive, data-driven approaches. Additionally, extending our joint-level acceleration modeling to subgroup or limb-level dynamics may further enrich the model’s understanding of human motion. We believe these directions, building on the foundation established by FineTec, will drive continued progress in robust action recognition.

\section{Conclusion}

In this work, we tackle the challenging problem of fine-grained action recognition under temporal corruption. 
We introduce FineTec, a unified framework specifically designed to address this issue.
FineTec first restores temporal continuity from corrupted inputs via a Context-aware Sequence Completion module. 
It then employs Skeleton-based Spatial Decomposition, guided by biological priors, to partition the skeleton and amplify subtle motion distinctions. 
Finally, a Physics-driven Acceleration Modeling module leverages Lagrangian dynamics to capture discriminative motion cues beyond simple displacement. 
Extensive experiments on both coarse-grained and fine-grained benchmarks demonstrate that FineTec consistently outperforms existing methods under varying degrees of temporal corruption. 
Future directions involve expanding our framework to multi-modal contexts and enhancing biomechanical modeling.
\section{Acknowledgments}
This work was funded by the National Natural Science Foundation of China (NSFC) under Grant 62306239, and supported by the Sanqin Talents Introduction Plan of Shaanxi Province, China.

\bibliography{aaai2026}

@inproceedings{shao2020FineGym,
  title={Finegym: A hierarchical video dataset for fine-grained action understanding},
  author={Shao, Dian and Zhao, Yue and Dai, Bo and Lin, Dahua},
  booktitle={Proceedings of the IEEE/CVF conference on computer vision and pattern recognition},
  pages={2616--2625},
  year={2020}
}

@article{soomro2012UCF101,
  title={UCF101: A Dataset of 101 Human Actions Classes From Videos in The Wild},
  author={ Soomro, Khurram  and  Zamir, Amir Roshan  and  Shah, Mubarak },
  journal={Computer Science},
  year={2012},
}

@inproceedings{yan2018ST-GCN,
  title={Spatial temporal graph convolutional networks for skeleton-based action recognition},
  author={Yan, Sijie and Xiong, Yuanjun and Lin, Dahua},
  booktitle={Proceedings of the AAAI conference on artificial intelligence},
  volume={32},
  number={1},
  year={2018}
}

@inproceedings{duan2022pyskl,
  title={Pyskl: Towards good practices for skeleton action recognition},
  author={Duan, Haodong and Wang, Jiaqi and Chen, Kai and Lin, Dahua},
  booktitle={Proceedings of the 30th ACM International Conference on Multimedia},
  pages={7351--7354},
  year={2022}
}

@inproceedings{lin2023actionlet,
  title={Actionlet-dependent contrastive learning for unsupervised skeleton-based action recognition},
  author={Lin, Lilang and Zhang, Jiahang and Liu, Jiaying},
  booktitle={Proceedings of the IEEE/CVF Conference on Computer Vision and Pattern Recognition},
  pages={2363--2372},
  year={2023}
}

@ARTICLE{chi2024InfoGCN++,
author={Chi, Seunggeun and Chi, Hyung-Gun and Huang, Qixing and Ramani, Karthik},
journal={ IEEE Transactions on Pattern Analysis \& Machine Intelligence },
title={{ InfoGCN++: Learning Representation by Predicting the Future for Online Skeleton-Based Action Recognition }},
year={2025},
volume={47},
number={01},
ISSN={1939-3539},
pages={514-528},
doi={10.1109/TPAMI.2024.3466212},
publisher={IEEE Computer Society},
address={Los Alamitos, CA, USA},
month=jan}

@inproceedings{zhou2024blockgcn,
  title={BlockGCN: Redefining Topology Awareness for Skeleton-Based Action Recognition},
  author={Zhou, Yuxuan and Yan, Xudong and Cheng, Zhi-Qi and Yan, Yan and Dai, Qi and Hua, Xian-Sheng},
  booktitle={Proceedings of the IEEE/CVF Conference on Computer Vision and Pattern Recognition},
  year={2024}
}

@misc{liu2023multidimensional,
      title={Multi-Dimensional Refinement Graph Convolutional Network with Robust Decouple Loss for Fine-Grained Skeleton-Based Action Recognition}, 
      author={Sheng-Lan Liu and Yu-Ning Ding and Jin-Rong Zhang and Kai-Yuan Liu and Si-Fan Zhang and Fei-Long Wang and Gao Huang},
      year={2023},
      eprint={2306.15321},
      archivePrefix={arXiv},
      primaryClass={cs.CV}
}

@INPROCEEDINGS{zhang2024pgvt,
  author={Zhang, Haosong and Leong, Mei Chee and Li, Liyuan and Lin, Weisi},
  booktitle={2024 IEEE/CVF Winter Conference on Applications of Computer Vision (WACV)}, 
  title={PGVT: Pose-Guided Video Transformer for Fine-Grained Action Recognition}, 
  year={2024},
  volume={},
  number={},
  pages={6631-6642},
  doi={10.1109/WACV57701.2024.00651}}

@InProceedings{andriluka2025learned,
author="Andriluka, Mykhaylo
and Tabanpour, Baruch
and Freeman, C. Daniel
and Sminchisescu, Cristian",
editor="Leonardis, Ale{\v{s}}
and Ricci, Elisa
and Roth, Stefan
and Russakovsky, Olga
and Sattler, Torsten
and Varol, G{\"u}l",
title="Learned Neural Physics Simulation for Articulated 3D Human Pose Reconstruction",
booktitle="Computer Vision -- ECCV 2024",
year="2025",
publisher="Springer Nature Switzerland",
address="Cham",
pages="320--336",
isbn="978-3-031-72907-2"
}

@ARTICLE{zhang2022pimnet,
  author={Zhang, Zhibo and Zhu, Yanjun and Rai, Rahul and Doermann, David},
  journal={IEEE Robotics and Automation Letters}, 
  title={PIMNet: Physics-Infused Neural Network for Human Motion Prediction}, 
  year={2022},
  volume={7},
  number={4},
  pages={8949-8955},
  doi={10.1109/LRA.2022.3188892}}

@article{duong23porthamiltonian,
author = {Thai Duong, Abdullah Altawaitan, Jason Stanley AND Nikolay Atanasov},
title = {Port-{H}amiltonian Neural {ODE} Networks on Lie Groups For Robot Dynamics Learning and Control},
journal = {arXiv preprint arXiv:2401.09520},
year = {2023},
}

@InProceedings{zhang2024incorporating,
    author    = {Zhang, Yufei and Kephart, Jeffrey O. and Ji, Qiang},
    title     = {Incorporating Physics Principles for Precise Human Motion Prediction},
    booktitle = {Proceedings of the IEEE/CVF Winter Conference on Applications of Computer Vision (WACV)},
    month     = {January},
    year      = {2024},
    pages     = {6164-6174}
}

@article{ionescu2014human36m,
  author = {Ionescu, Catalin and Papava, Dragos and Olaru, Vlad and Sminchisescu,  Cristian},
  title = {Human3.6M: Large Scale Datasets and Predictive Methods for 3D Human Sensing in Natural Environments},
  journal = {IEEE Transactions on Pattern Analysis and Machine Intelligence},
  publisher = {IEEE Computer Society},
  volume = {36},
  number = {7},
  pages = {1325-1339},
  month = {jul},
  year = {2014}
}

@inproceedings{ye2022ostrack,
  title={Joint Feature Learning and Relation Modeling for Tracking: A One-Stream Framework},
  author={Ye, Botao and Chang, Hong and Ma, Bingpeng and Shan, Shiguang and Chen, Xilin},
  booktitle={ECCV},
  year={2022}
}

@inproceedings{SunXLW19,
  title={Deep High-Resolution Representation Learning for Human Pose Estimation},
  author={Ke Sun and Bin Xiao and Dong Liu and Jingdong Wang},
  booktitle={CVPR},
  year={2019}
}

@article{jain1995diagonalized,
  title={Diagonalized Lagrangian robot dynamics},
  author={Jain, Abhinandan and Rodriguez, Guillermo},
  journal={IEEE Transactions on Robotics and Automation},
  volume={11},
  number={4},
  pages={571--584},
  year={1995},
  publisher={IEEE}
}

@inproceedings{shahroudy2016ntu,
  title={NTU RGB+D: A large scale dataset for 3D human activity analysis},
  author={Shahroudy, Amir and Liu, Jun and Ng, Tian-Tsong and Wang, Gang},
  booktitle={CVPR},
  pages={1010--1019},
  year={2016}
}

@article{liu2020ntu,
  title={NTU RGB+D 120: A large-scale benchmark for 3D human activity understanding},
  author={Liu, Jun and Shahroudy, Amir and Perez, Mauricio and Wang, Gang and Duan, Ling-Yu and Kot, Alex C},
  journal={IEEE Transactions on Pattern Analysis and Machine Intelligence},
  volume={42},
  number={10},
  pages={2684--2701},
  year={2020}
}

@article{shi2020skeleton,
  title={Skeleton-based action recognition with multi-stream adaptive graph convolutional networks},
  author={Shi, Lei and Zhang, Yifan and Cheng, Jian and Lu, Hanqing},
  journal={IEEE Transactions on Image Processing},
  volume={29},
  pages={9532--9545},
  year={2020},
  publisher={IEEE}
}

@article{huang2025sefar,
  title={SeFAR: Semi-supervised Fine-grained Action Recognition with Temporal Perturbation and Learning Stabilization},
  author={Huang, Yongle and Chen, Haodong and Xu, Zhenbang and Jia, Zihan and Sun, Haozhou and Shao, Dian},
  journal={arXiv preprint arXiv:2501.01245},
  year={2025}
}

@inproceedings{chen2021channel,
  title={Channel-wise topology refinement graph convolution for skeleton-based action recognition},
  author={Chen, Yuxin and Zhang, Ziqi and Yuan, Chunfeng and Li, Bing and Deng, Ying and Hu, Weiming},
  booktitle={Proceedings of the IEEE/CVF international conference on computer vision},
  pages={13359--13368},
  year={2021}
}

@inproceedings{yang2020temporal,
  title={Temporal pyramid network for action recognition},
  author={Yang, Ceyuan and Xu, Yinghao and Shi, Jianping and Dai, Bo and Zhou, Bolei},
  booktitle={Proceedings of the IEEE/CVF conference on computer vision and pattern recognition},
  pages={591--600},
  year={2020}
}

@article{wang2018temporal,
  title={Temporal segment networks for action recognition in videos},
  author={Wang, Limin and Xiong, Yuanjun and Wang, Zhe and Qiao, Yu and Lin, Dahua and Tang, Xiaoou and Van Gool, Luc},
  journal={IEEE transactions on pattern analysis and machine intelligence},
  volume={41},
  number={11},
  pages={2740--2755},
  year={2018},
  publisher={IEEE}
}

@article{lin2025exploring,
  title={Exploring the evolution of physics cognition in video generation: A survey},
  author={Lin, Minghui and Wang, Xiang and Wang, Yishan and Wang, Shu and Dai, Fengqi and Ding, Pengxiang and Wang, Cunxiang and Zuo, Zhengrong and Sang, Nong and Huang, Siteng and others},
  journal={arXiv preprint arXiv:2503.21765},
  year={2025}
}

@inproceedings{gartner2022differentiable,
  title={Differentiable dynamics for articulated 3d human motion reconstruction},
  author={G{\"a}rtner, Erik and Andriluka, Mykhaylo and Coumans, Erwin and Sminchisescu, Cristian},
  booktitle={Proceedings of the IEEE/CVF conference on computer vision and pattern recognition},
  pages={13190--13200},
  year={2022}
}

@inproceedings{gartner2022trajectory,
  title={Trajectory optimization for physics-based reconstruction of 3d human pose from monocular video},
  author={G{\"a}rtner, Erik and Andriluka, Mykhaylo and Xu, Hongyi and Sminchisescu, Cristian},
  booktitle={Proceedings of the IEEE/CVF Conference on Computer Vision and Pattern Recognition},
  pages={13106--13115},
  year={2022}
}

@inproceedings{wang2024skeleton,
  title={Skeleton-in-context: Unified skeleton sequence modeling with in-context learning},
  author={Wang, Xinshun and Fang, Zhongbin and Li, Xia and Li, Xiangtai and Chen, Chen and Liu, Mengyuan},
  booktitle={Proceedings of the IEEE/CVF Conference on Computer Vision and Pattern Recognition},
  pages={2436--2446},
  year={2024}
}

@inproceedings{xie2025spatial,
  title={Are Spatial-Temporal Graph Convolution Networks for Human Action Recognition Over-Parameterized?},
  author={Xie, Jianyang and Zhao, Yitian and Meng, Yanda and Zhao, He and Nguyen, Anh and Zheng, Yalin},
  booktitle={CVPR},
  pages={24309--24319},
  year={2025}
}

@inproceedings{zheng2024nettrack,
  title={Nettrack: Tracking highly dynamic objects with a net},
  author={Zheng, Guangze and Lin, Shijie and Zuo, Haobo and Fu, Changhong and Pan, Jia},
  booktitle={Proceedings of the IEEE/CVF Conference on Computer Vision and Pattern Recognition},
  pages={19145--19155},
  year={2024}
}

@inproceedings{zhu2025semantic,
  title={Semantic-guided Cross-Modal Prompt Learning for Skeleton-based Zero-shot Action Recognition},
  author={Zhu, Anqi and Zhu, Jingmin and Bailey, James and Gong, Mingming and Ke, Qiuhong},
  booktitle={Proceedings of the Computer Vision and Pattern Recognition Conference},
  pages={13876--13885},
  year={2025}
}

@inproceedings{wang2025action,
  title={Action Detail Matters: Refining Video Recognition with Local Action Queries},
  author={Wang, Mengmeng and Huang, Zeyi and Kong, Xiangjie and Shen, Guojiang and Dai, Guang and Wang, Jingdong and Liu, Yong},
  booktitle={Proceedings of the Computer Vision and Pattern Recognition Conference},
  pages={19132--19142},
  year={2025}
}

@inproceedings{li2022dynamic,
  title={Dynamic spatio-temporal specialization learning for fine-grained action recognition},
  author={Li, Tianjiao and Foo, Lin Geng and Ke, Qiuhong and Rahmani, Hossein and Wang, Anran and Wang, Jinghua and Liu, Jun},
  booktitle={European Conference on Computer Vision},
  pages={386--403},
  year={2022},
  organization={Springer}
}

@inproceedings{xie2024dynamic,
  title={Dynamic semantic-based spatial graph convolution network for skeleton-based human action recognition},
  author={Xie, Jianyang and Meng, Yanda and Zhao, Yitian and Nguyen, Anh and Yang, Xiaoyun and Zheng, Yalin},
  booktitle={Proceedings of the AAAI conference on artificial intelligence},
  volume={38},
  number={6},
  pages={6225--6233},
  year={2024}
}

@article{myung2024degcn,
  title={Degcn: Deformable graph convolutional networks for skeleton-based action recognition},
  author={Myung, Woomin and Su, Nan and Xue, Jing-Hao and Wang, Guijin},
  journal={IEEE Transactions on Image Processing},
  volume={33},
  pages={2477--2490},
  year={2024},
  publisher={IEEE}
}

@inproceedings{liu2025revealing,
  title={Revealing key details to see differences: A novel prototypical perspective for skeleton-based action recognition},
  author={Liu, Hongda and Liu, Yunfan and Ren, Min and Wang, Hao and Wang, Yunlong and Sun, Zhenan},
  booktitle={Proceedings of the Computer Vision and Pattern Recognition Conference},
  pages={29248--29257},
  year={2025}
}

@article{jiang2024lighter,
  title={Lighter and faster: A multi-scale adaptive graph convolutional network for skeleton-based action recognition},
  author={Jiang, Yuanjian and Deng, Hongmin},
  journal={Engineering Applications of Artificial Intelligence},
  volume={132},
  pages={107957},
  year={2024},
  publisher={Elsevier}
}

@article{leong2022combined,
  title={Combined CNN transformer encoder for enhanced fine-grained human action recognition},
  author={Leong, Mei Chee and Zhang, Haosong and Tan, Hui Li and Li, Liyuan and Lim, Joo Hwee},
  journal={arXiv preprint arXiv:2208.01897},
  year={2022}
}

@inproceedings{kim2025videoicl,
  title={VideoICL: Confidence-based Iterative In-context Learning for Out-of-Distribution Video Understanding},
  author={Kim, Kangsan and Park, Geon and Lee, Youngwan and Yeo, Woongyeong and Hwang, Sung Ju},
  booktitle={Proceedings of the Computer Vision and Pattern Recognition Conference},
  pages={3295--3305},
  year={2025}
}

@inproceedings{zhang2021temporal,
  title={Temporal query networks for fine-grained video understanding},
  author={Zhang, Chuhan and Gupta, Ankush and Zisserman, Andrew},
  booktitle={Proceedings of the ieee/cvf conference on computer vision and pattern recognition},
  pages={4486--4496},
  year={2021}
}

@inproceedings{liu2024physgen,
  title={Physgen: Rigid-body physics-grounded image-to-video generation},
  author={Liu, Shaowei and Ren, Zhongzheng and Gupta, Saurabh and Wang, Shenlong},
  booktitle={European Conference on Computer Vision},
  pages={360--378},
  year={2024},
  organization={Springer}
}

@inproceedings{chi2022infogcn,
  title={Infogcn: Representation learning for human skeleton-based action recognition},
  author={Chi, Hyung-gun and Ha, Myoung Hoon and Chi, Seunggeun and Lee, Sang Wan and Huang, Qixing and Ramani, Karthik},
  booktitle={Proceedings of the IEEE/CVF conference on computer vision and pattern recognition},
  pages={20186--20196},
  year={2022}
}

@inproceedings{zhang2024physpt,
  title={Physpt: Physics-aware pretrained transformer for estimating human dynamics from monocular videos},
  author={Zhang, Yufei and Kephart, Jeffrey O and Cui, Zijun and Ji, Qiang},
  booktitle={Proceedings of the IEEE/CVF Conference on Computer Vision and Pattern Recognition},
  pages={2305--2317},
  year={2024}
}

@inproceedings{ugrinovic2024multiphys,
  title={Multiphys: Multi-person physics-aware 3d motion estimation},
  author={Ugrinovic, Nicolas and Pan, Boxiao and Pavlakos, Georgios and Paschalidou, Despoina and Shen, Bokui and Sanchez-Riera, Jordi and Moreno-Noguer, Francesc and Guibas, Leonidas},
  booktitle={Proceedings of the IEEE/CVF Conference on Computer Vision and Pattern Recognition},
  pages={2331--2340},
  year={2024}
}

@inproceedings{FineQuest,
author = {Chen, Haodong and Huang, Haojian and Yin, Xinxiang and Shao, Dian},
title = {FineQuest: Adaptive Knowledge-Assisted Sports Video Understanding via Agent-of-Thoughts Reasoning},
year = {2025},
isbn = {9798400720352},
publisher = {Association for Computing Machinery},
address = {New York, NY, USA},
doi = {10.1145/3746027.3754885},
booktitle = {Proceedings of the 33rd ACM International Conference on Multimedia},
pages = {2909–2918},
numpages = {10},
location = {Dublin, Ireland},
series = {MM '25}
}

@inproceedings{shao2020intra,
  title={Intra-and inter-action understanding via temporal action parsing},
  author={Shao, Dian and Zhao, Yue and Dai, Bo and Lin, Dahua},
  booktitle={Proceedings of the IEEE/CVF conference on computer vision and pattern recognition},
  pages={730--739},
  year={2020}
}

@InProceedings{Shao_2025_CVPR,
    author    = {Shao, Dian and Shi, Mingfei and Xu, Shengda and Chen, Haodong and Huang, Yongle and Wang, Binglu},
    title     = {FinePhys: Fine-grained Human Action Generation by Explicitly Incorporating Physical Laws for Effective Skeletal Guidance},
    booktitle = {Proceedings of the IEEE/CVF Conference on Computer Vision and Pattern Recognition},
    month     = {June},
    year      = {2025},
    pages     = {1905-1916}
}

@article{rajendran2024review,
  title={Review on synergizing the Metaverse and AI-driven synthetic data: enhancing virtual realms and activity recognition in computer vision},
  author={Rajendran, Megani and Tan, Chek Tien and Atmosukarto, Indriyati and Ng, Aik Beng and See, Simon},
  journal={Visual Intelligence},
  volume={2},
  number={1},
  pages={27},
  year={2024},
  publisher={Springer}
}

\clearpage

\appendix
\section{Training \& Dataset Details}
\subsection{Overview}
We deploy FineTec using PyTorch, and the training process consists of two steps: \ding{182} Pre-train the Context-aware Sequence Completion module on skeletal datasets;
\ding{183} Train the FineTec framework, while keeping the Completion module frozen.

For the initial pre-training phase of the Context-aware Sequence Completion module on skeletal datasets, the training regimen in Gym99 and Gym288-skeleton datasets is established as follows. 
The model processes input sequences of 16 frames. 
Each frame's representation is a concatenation of 34-dimensional skeletal features, a 1-dimensional mask indicator, and 8-dimensional positional encodings, yielding a 43-dimensional input per frame. 
Meanwhile, skeletons from Human3.6M datasets~\cite{ionescu2014human36m} are utilized to form both query-prior and in-context pairs, which are processed into the backbone.
The core network architecture is a 48-layer S-MLPs and T-MLPs incorporating normalization. 
An input fully connected layer maps the 43-dimensional frame-wise input to an internal embedding dimension of 34, employing a ReLU activation. 
Subsequently, an output fully connected layer projects the processed embeddings back to the 34-dimensional skeletal feature space, also utilizing ReLU activation, and its weights are initialized using a truncated normal distribution.

For the second training stage, where FineTec is trained while the parameters of the pre-trained Context-aware Sequence Completion module are kept frozen, we configure the training process as follows. 
Optimization is conducted using the Stochastic Gradient Descent (SGD) algorithm, incorporating Nesterov momentum. 
The optimizer is initialized with a learning rate from 0.05 to 0.2, a momentum coefficient of 0.9, and an L2 weight decay of ($5 \times 10^{-4}$). 
To dynamically adjust the learning rate, a cosine annealing schedule is applied over 200 epochs, reducing the learning rate to a minimum of 0. 

In terms of both NTU-60-xsub~\cite{shahroudy2016ntu} and NTU-120-xsub~\cite{liu2020ntu} skeleton datasets, the input sequence length is set to 100 frames. 
And skeletons in Human3.6M are also sampled with 100 frames.

\begin{figure}[t]
    \centering
    \includegraphics[width=1.0\linewidth]{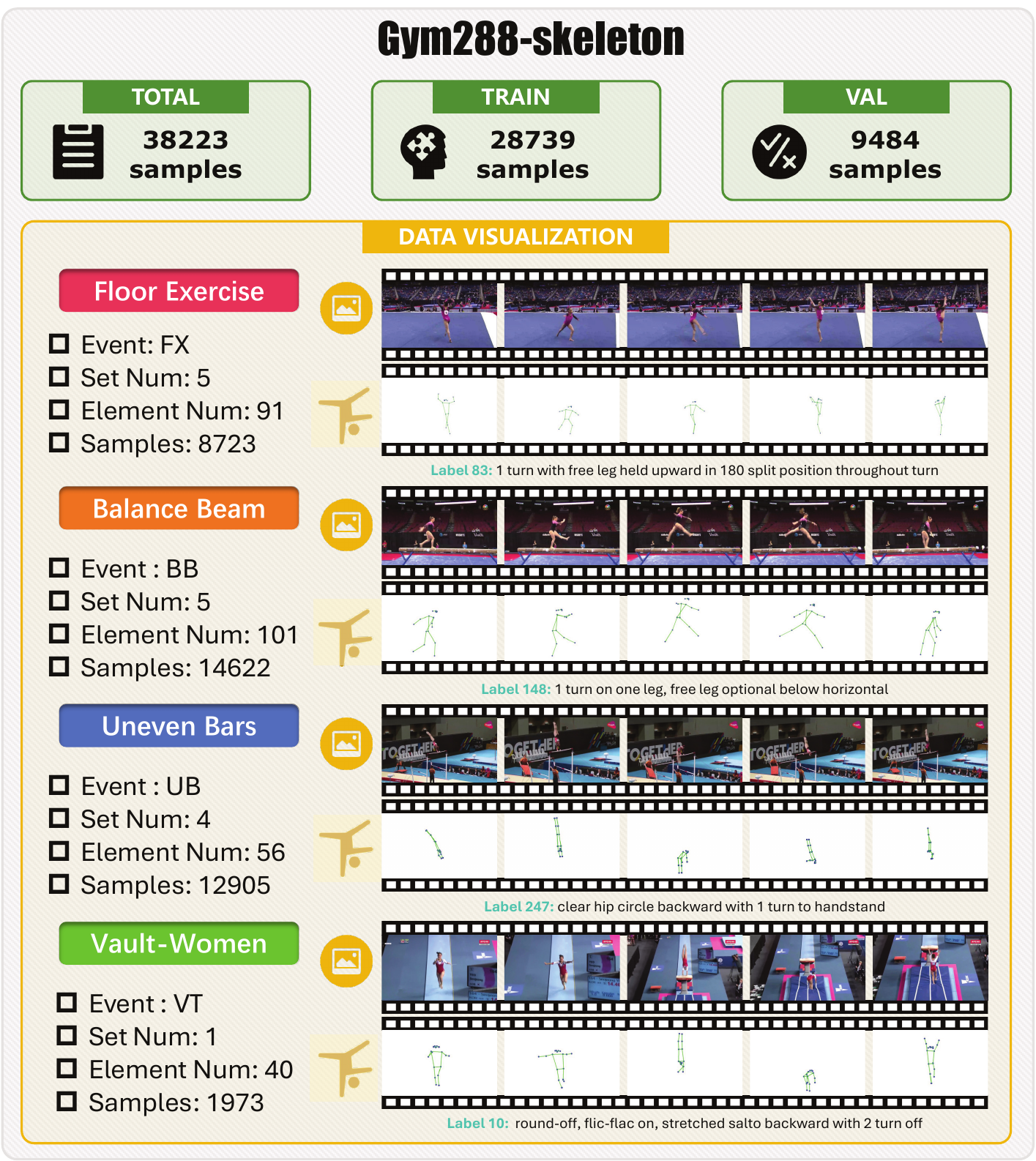}
    \caption{\textbf{Statistics and Visualized Analysis of our purposed Gym288-skeleton Dataset.} We present representative examples and a statistical comparison with the Gym99 Skeleton Dataset.
    }
    \label{supp_fig:appendix_gym288_visualization}
\end{figure}

\subsection{Context-aware Sequence Completion Module Pre-training}

The training is performed with a batch size of 256, supported by 8 data loader workers. 
Optimization is carried out using an Adam optimizer, governed by a cosine annealing learning rate schedule. 
The learning rate cycles between a maximum of ($1 \times 10^{-5}$) and a minimum of ($5 \times 10^{-8}$) over a total of 40,000 iterations. 
A weight decay of ($1 \times 10^{-4}$) is applied for regularization. 
To enhance model robustness, data augmentation techniques are utilized. 
Furthermore, the model is configured to process derivative information from both input and output signals, potentially capturing velocity or acceleration dynamics. 
All experiments are conducted with a fixed random seed of 304 to ensure reproducibility.

\subsection{FineTec Training}
During training, we use a batch size of 64. 
Input 2D skeletal joint data is processed through a pipeline involving pre-normalization and feature generation, followed by uniform sampling of sequences into 16-frame clips.
The overall training duration is set for a maximum of 150 epochs. 
And this training step of Gym288-skeleton is conducted on a Linux (Ubuntu) machine with 4 NVIDIA 4090 GPUs within 5 hours.

\begingroup
\setlength{\tabcolsep}{2pt}
\renewcommand{\arraystretch}{1.2}
\footnotesize
\begin{table}[t]
  \centering
  \caption{\textbf{Comparison with state-of-the-art methods on UCF101.} Three levels of temporal corruption were set: minor, moderate, and severe. The evaluation metrics include top-1 accuracy and top-5 accuracy.}
  \begin{tabular}{lcccccc}
    \toprule
    & \multicolumn{2}{c}{\textbf{UCF-Min.}} & \multicolumn{2}{c}{\textbf{UCF-Mod.}} & \multicolumn{2}{c}{\textbf{UCF-Sev.}} \\
    \cmidrule(lr){2-3}\cmidrule(lr){4-5}\cmidrule(lr){6-7}
    \multirow{-3}{*}{\textbf{Method}}  & \textbf{Top-1} & \textbf{Top-5} & \textbf{Top-1} & \textbf{Top-5} & \textbf{Top-1} & \textbf{Top-5} \\
    \midrule
    ST-GCN $\textit{\scriptsize{\textcolor[RGB]{130,130,130}{AAAI'18}}}$  & 0.648 & 0.852 & 0.632 & 0.858 & 0.582 & 0.844 \\
    AAGCN $\textit{\scriptsize{\textcolor[RGB]{130,130,130}{TIP'20}}}$  & 0.634 & 0.851 & 0.636 & 0.861 & 0.638 & 0.843 \\
    \midrule
    \rowcolor{cyan!10}
    \textbf{FineTec (Ours)} & \textbf{0.652} & \textbf{0.861} & \textbf{0.638} & \textbf{0.859} & \textbf{0.621} & \textbf{0.847} \\
    \bottomrule
  \end{tabular}
  \label{tab:appendix_results_ucf101}
\end{table}
\endgroup

\section{Model Details}

\begin{figure*}[t]
    \centering
    \includegraphics[width=1.0\linewidth]{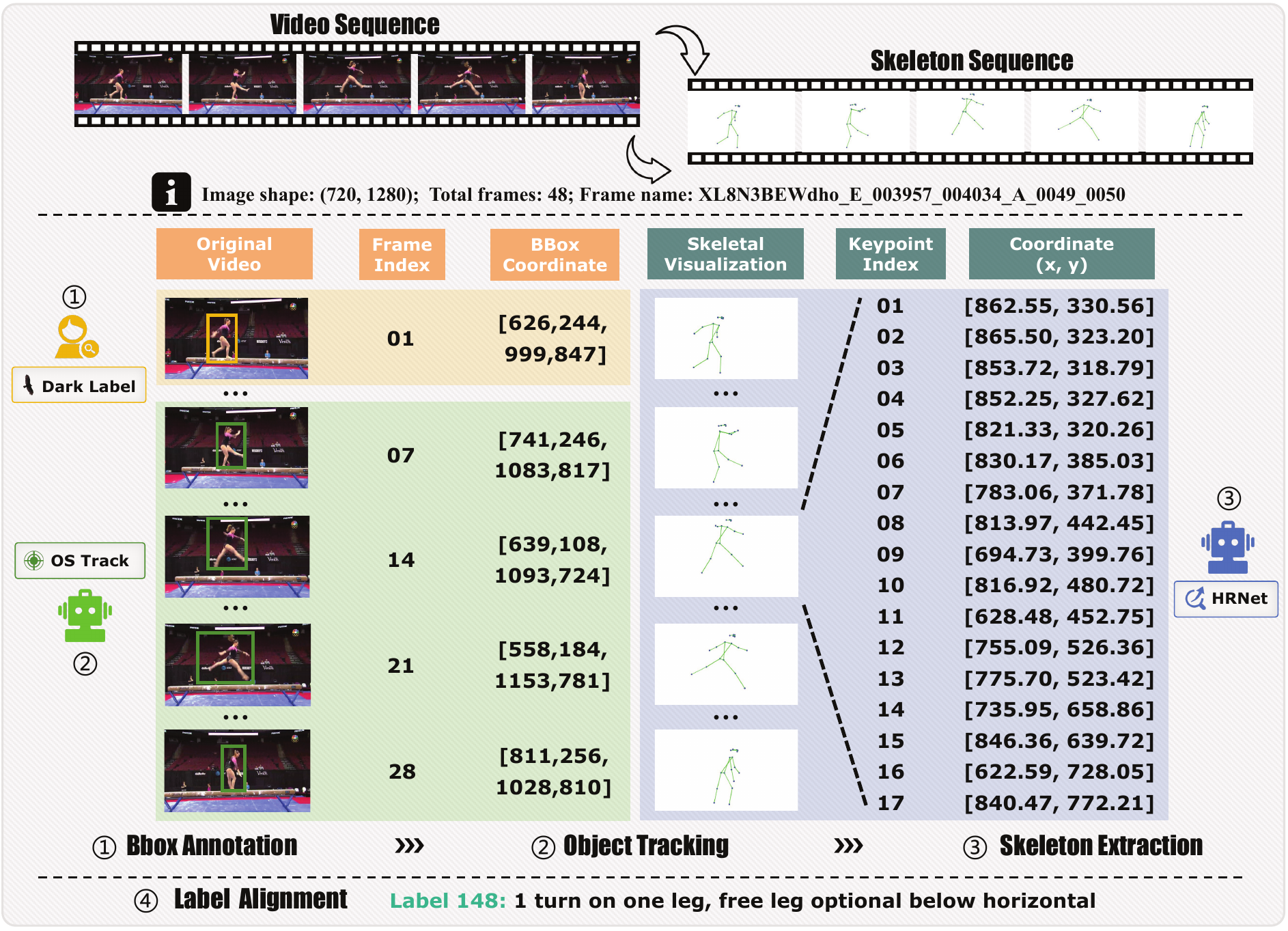}
    \caption{\textbf{Annotation Process of the Gym288 Skeleton Dataset on Label 148.} The skeleton comes from the elemental class \textit{``1 turn on one leg, free leg optional below horizontal"}, the set class \textit{``switch leap (leap forward with leg change)''} and the event class \textit{``Balance Beam''}.
    }
    \label{supp_fig:fig_annotation_process_label_148}
\end{figure*}

\subsection{Mask Strategies in the Context-aware Sequence Completion Module}
To enhance robustness of FineTec and its ability to generalize across various data corruption types, it employs five distinct masking strategies on the reference sequence: 
1) Random masking, which simulates sporadic data loss by dropping arbitrary frames; 
2) Pattern-based masking, which replicates the specific mask pattern of the input query to train for structured gaps; 
3) Left-side masking, which removes a continuous block from the beginning, analogous to a delayed motion capture start; 
4) Right-side masking, which removes a block from the end, simulating early termination or forecasting tasks; 
and 5) Middle masking, which removes a central block to represent occlusions during an action. 
This diverse set of strategies ensures the Completion module is trained on a wide range of missing data scenarios, significantly improving its performance and reliability for real-world motion in-painting tasks.

\subsection{Augmentation Techniques in the Skeleton-based Spatial Decomposition Module}

\begin{figure}[t]
    \centering
    \includegraphics[width=1.0\linewidth]{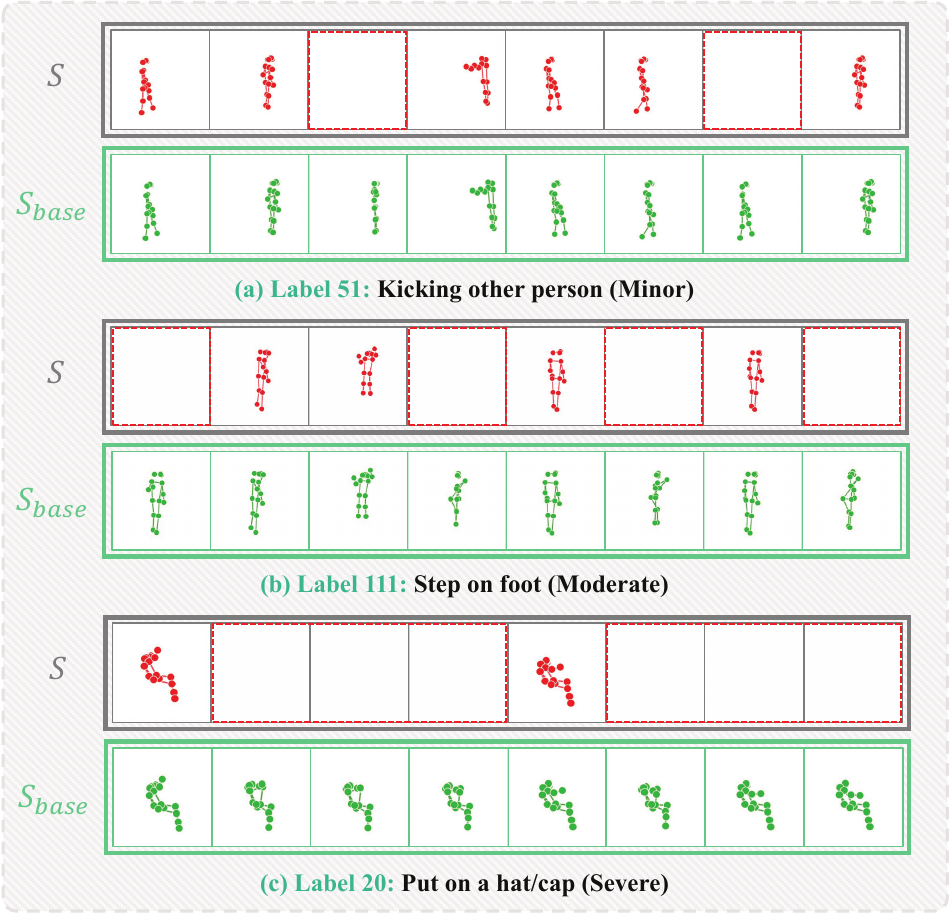}
    \caption{\textbf{Visualization of restoration on NTU-120-xsub.} Our Completion module produces complete and consistent skeletons using temporal-corrupted skeleton sequences.
    }
    \label{fig:appendix_ntu_restoration_visualization}
\end{figure}

To enhance the ability of FineTec to capture fine-grained motion details, we employ both weak and strong augmentation strategies. 
Weak augmentation involves only temporal cropping to create slight temporal variations. 
The extreme augmentation pipeline, however, applies a diverse set of transformations to generate more challenging samples. 
These transformations can be broadly classified into three categories: 
(\textit{i}) \textit{Spatial augmentations}, which alter the skeletal structure within each frame through operations like random spatial flipping, Gaussian noise, axis masking, and bone rescaling; 
(\textit{ii}) \textit{Temporal augmentations}, which manipulate the sequence's dynamics via random time flipping and interpolation; 
and (\textit{iii}) \textit{Spatio-temporal augmentations}, such as dropout, that introduce random occlusions across both space and time. 
This comprehensive strategy is designed to force FineTec to learn robust features that are invariant to significant appearance and motion variations, thereby enhancing its ability to distinguish between dynamic and static body parts.

\section{Gym288-skeleton Dataset}

To facilitate a more rigorous analysis, we have augmented existing skeleton datasets by introducing a novel dataset annotated on the FineGym~\cite{shao2020FineGym} benchmark. 
This dataset is constructed by extending the open-source Gym99 dataset available in MMAction2.
This section details our annotation pipeline, presents statistical analyses of the newly curated dataset, Gym288-skeleton, and includes illustrative visualization skeletons.

\subsection{Formulation Pipeline}

\begin{figure*}[t]
    \centering
    \includegraphics[width=1.0\linewidth]{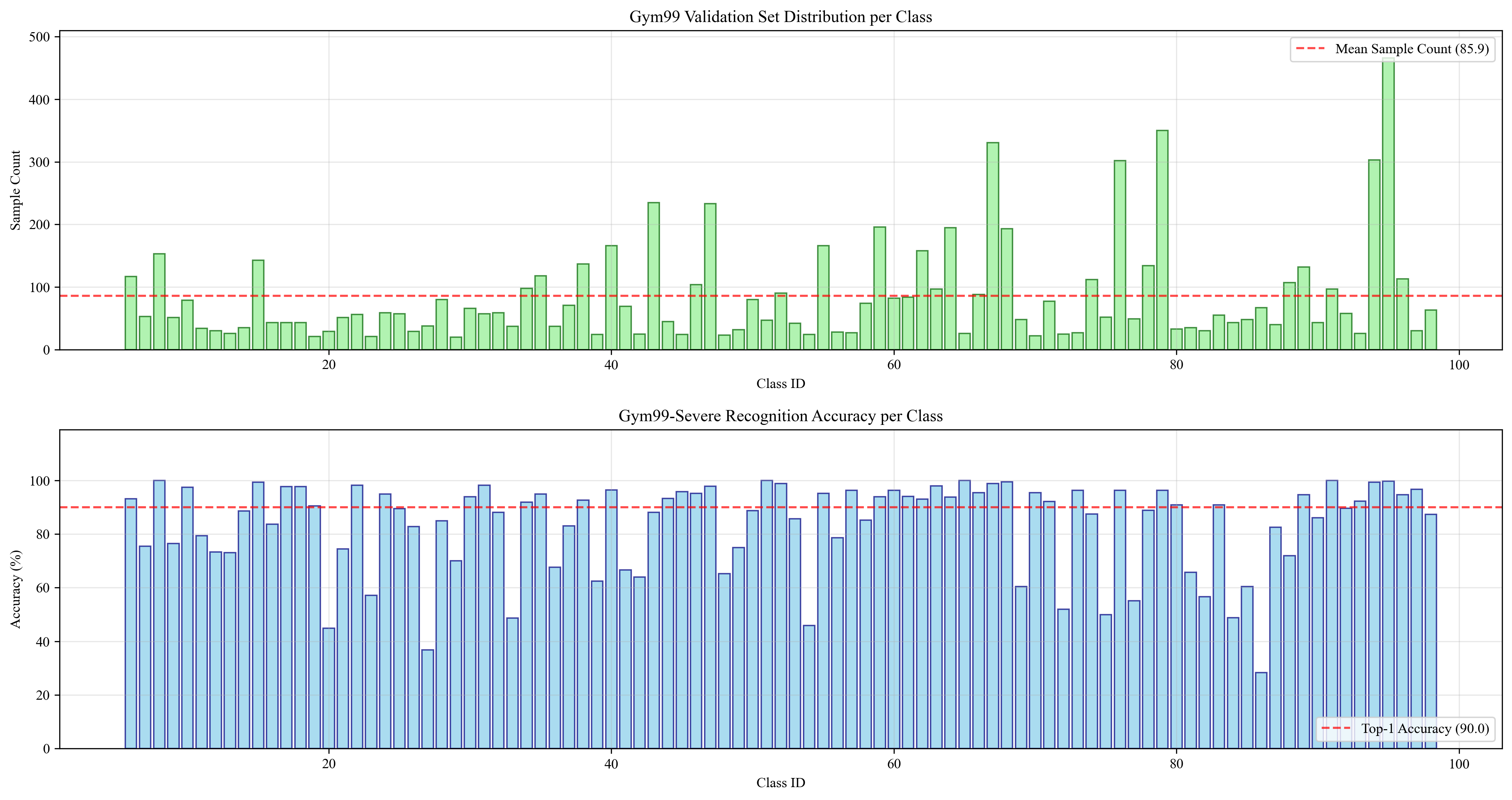}
    \caption{\textbf{Gym99-skeleton recognition Analysis}. The Top-1 accuracy is used. The upper figure illustrates the number of action sequences per class in the validation set, while the lower figure shows the corresponding per-class accuracy achieved by the FineTec model.
    }
    \label{fig:appendix_gym99_recognition_analysis}
\end{figure*}

To ensure accurate target identification for skeleton extraction, we perform manual annotation.
Specifically, bounding boxes for the primary athlete are manually annotated in the initial frame of 39,092 videos sourced from the FineGym dataset, which precisely define the targets for subsequent skeleton extraction.
Following the identification of the initial target, the OS Track algorithm~\cite{ye2022ostrack} is utilized to perform frame-by-frame object tracking throughout each video. 
This process yield the spatio-temporal trajectories of the primary athlete and generate the structured annotation files including frame-level skeletonal coordinate information.
With the HRNet skeleton annotator~\cite{SunXLW19} and the above bounding boxes, we extract 17 human skeletal pose keypoints within each frame.
And after data cleaning, we remove invalid or inaccurately identified data and curate a dataset comprising 38,223 skeletal action sequences of the primary subjects successfully.
Finally, to determine the action category for each sequence, we match the video filenames with the Gym288 classification index to assign a specific action label to each skeletal sequence, completing the creation of the Gym288-skeleton dataset.

\subsection{Data Statistics and Visualization}

\begin{figure*}[t]
    \centering
    \includegraphics[width=1.0\linewidth]{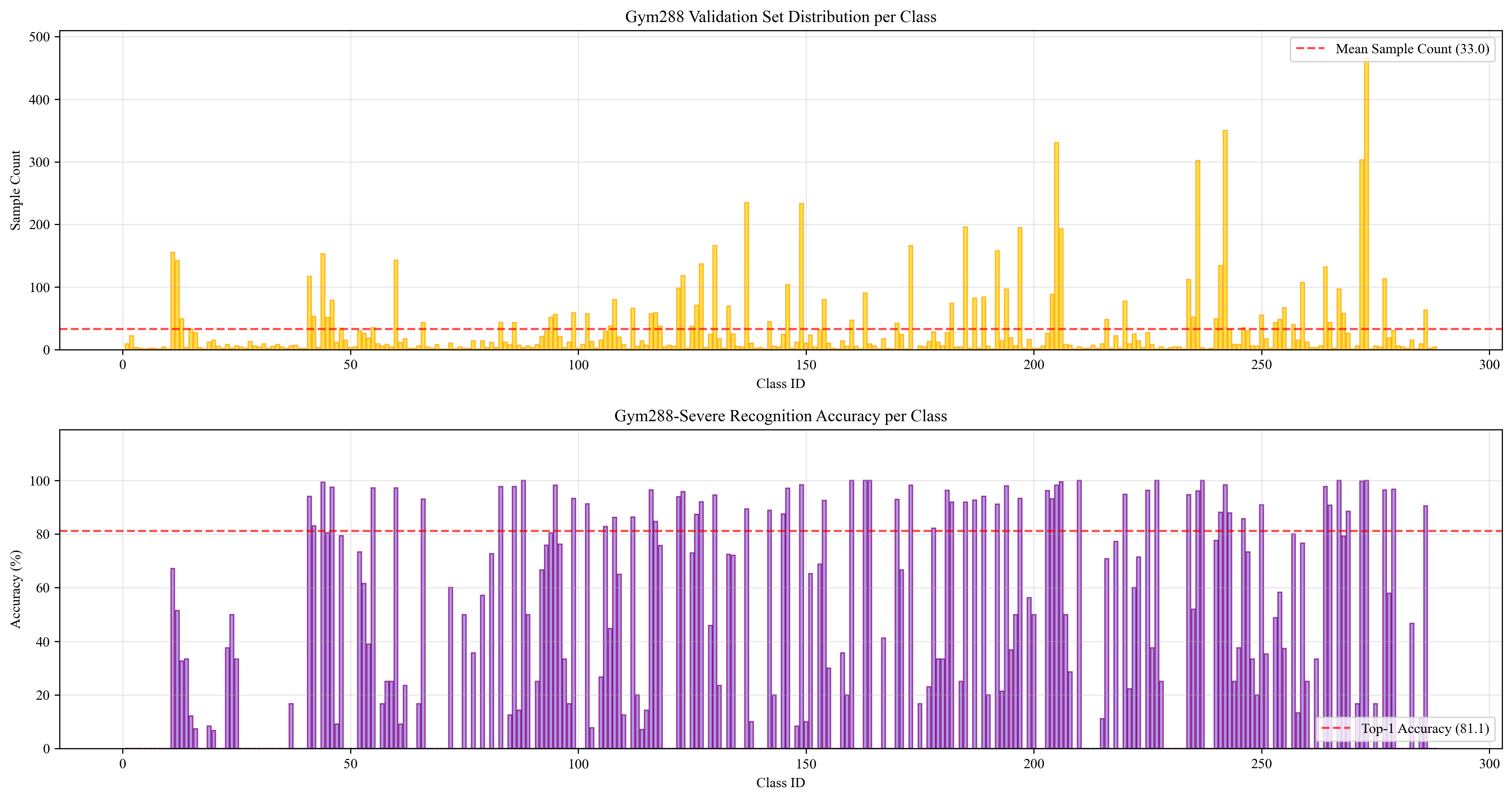}
    \caption{\textbf{Gym288-skeleton datasets recognition Analysis}. The Top-1 accuracy is used. The upper figure illustrates the number of action sequences per class in the validation set, while the lower figure shows the corresponding per-class accuracy achieved by the FineTec model.
    }
    \label{fig:appendix_gym288_recognition_analysis}
\end{figure*}

Our Gym288 skeleton dataset contains a total of 38,223 skeleton sequence samples. 
These are categorized into four major disciplines: Balance Beam (BB), Floor Exercise (FX), Uneven Bars (UB), and Vault (VT), covering 288 fine-grained action classes, following FineGym~\cite{shao2020FineGym}.
The dataset is divided into 28,739 training samples and 9,484 validation samples. 
Gym288 is characterized by significant action diversity. 
For example, in Balance Beam, frequently occurring leap and turn movements include ``split jump'' (972 instances) and ``switch leap to ring position'' (350 instances). 
Similarly, in Uneven Bars, complex aerial movements on the bars such as ``giant circle backward'' (1,582 instances) and ``transition flight between bars'' (3,394 instances) are highly represented.
Meanwhile, the Gym99 skeleton dataset comprises 29,005 skeleton sequence samples. 
It also covers the four major disciplines but includes only 99 action classes, with 19,121 training and 7,986 validation samples. 


\section{Additional Illustration \& Analysis}

\subsection{Elaboration on Euler-Lagrange Equations}
In the main paper, we adopt the following standard form of the equations of motion in Lagrangian mechanics:
\begin{align}
    M(q_t)\ddot{q}_t + C(q_t, \dot{q}_t)\dot{q}_t + g(q_t) = \tau_t,
\end{align}
where $M(q_t)$ is the configuration-dependent inertia matrix characterizing the mass distribution of the system, $C(q_t, \dot{q}_t)\dot{q}_t$ encapsulates the Coriolis and centrifugal forces as functions of joint positions $q_t$ and velocities $\dot{q}_t$, $g(q_t)$ represents the gravitational forces dependent on the configuration $q_t$, and $\tau_t$ denotes the vector of generalized forces, including joint torques and external forces. This formulation is derived from the classical Euler-Lagrange equations,
\begin{align}
    \frac{d}{dt}\frac{\partial L}{\partial \dot{q}^i}(t, q(t), \dot{q}(t)) - \frac{\partial L}{\partial q^i}(t, q(t), \dot{q}(t)) = \tau^i_t,
\end{align}
where the Lagrangian $L = T - V$ is defined by the kinetic energy $T = \frac{1}{2} \dot{q}^T M(q) \dot{q}$ and the potential energy $V = V(q)$. By computing the relevant derivatives,
\begin{align}
    \frac{\partial L}{\partial q^i} &= -\frac{\partial V}{\partial q^i} + \frac{1}{2} \dot{q}^T \frac{\partial M(q)}{\partial q^i} \dot{q}, \\
    \frac{d}{dt} \frac{\partial L}{\partial \dot{q}^i} &= M_{ij}(q) \ddot{q}^j + \dot{q}^j \frac{\partial M_{ij}}{\partial q^k} \dot{q}^k,
\end{align}
and substituting into the Euler-Lagrange equation, we obtain
\begin{align}
    M(q)\ddot{q} + C(q, \dot{q})\dot{q} + g(q) = \tau,
\end{align}
where $C(q, \dot{q})\dot{q}$ collects the Coriolis and centrifugal terms, $g(q) = \frac{\partial V}{\partial q}$ is the gravitational force, and $\tau$ is the generalized force vector. This compact form is widely used in robotics and multibody dynamics for modeling and control.

\subsection{Visualization of Restoration on NTU-120-XSub}

Figure~\ref{fig:appendix_ntu_restoration_visualization} provides a qualitative analysis of skeleton restoration on the NTU-120-XSub dataset, a coarse-grained benchmark. 
Leveraging our Context-aware Sequence Completion module, we demonstrate the restoration of skeletal data under three corrupted scenarios across three representative action classes. 
For conciseness and due to page limitations, we visualize only 8 frames from the complete input sequence of 100 frames. 
The results on NTU-120 underscore the effectiveness of our module in achieving high-fidelity restoration across a diverse set of actions. 
This indicates our module's strong generalization ability, making it a promising approach for various action recognition scenarios, including more complex, fine-grained datasets like FineGym.
This robust performance underscores our capability to effectively reconstruct the spatio-temporal integrity of human movements, highlighting its strong generalization ability on datasets with varying levels of action granularity.

\begin{figure}[t]
    \centering
    \includegraphics[width=1.0\linewidth]{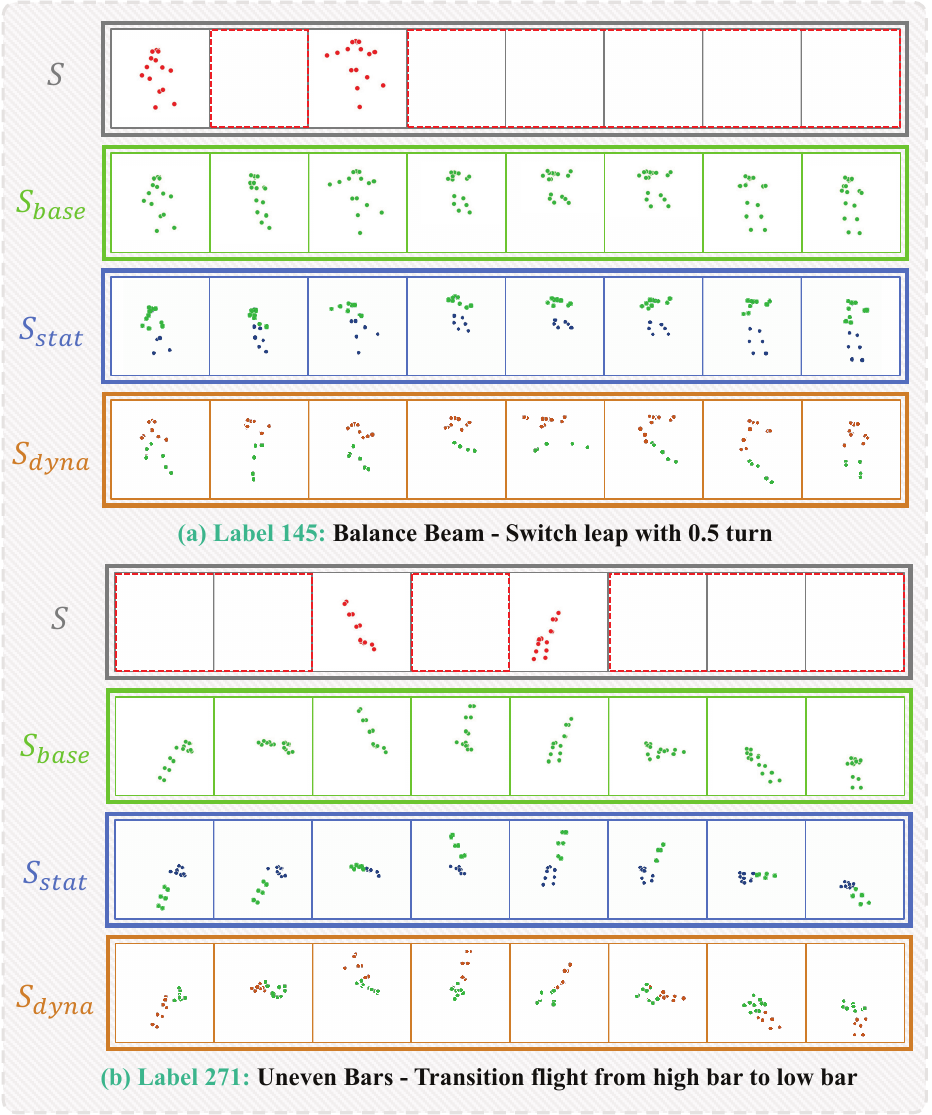}
    \caption{\textbf{Visualization of Three Components on the Gym288-skeleton Dataset in the Skeleton-based Spatial Decomposition.} Examples from three sets are visualized, with the temporal-corrupted skeleton inputs.
    }
    \label{fig:appendix_gym288_finesud_visualization}
\end{figure}

\subsection{Analysis of Three Components in the Skeleton-based Spatial Decomposition Module}

Figure~\ref{fig:appendix_gym288_finesud_visualization} visualizes the three component sequences in Decomposition module, alongside their corresponding input skeleton sequence. 
To better elucidate how it leverages biological priors for fine-grained analysis within the skeletal space, we select three distinct action sets from the Gym288-skeleton dataset and present one representative element class (\textit{i.e.}, label 43, 145, 271) from each for visualization. 
As can be observed, it effectively partitions the completed skeletons into DynamicUnits and StaticUnits, resulting in all sequences that are distinct from one another.
This decomposition isolates the most discriminative body parts for action recognition, thereby amplifying the subtle yet crucial distinctions necessary for fine-grained analysis.

\subsection{Experiment Results on the UCF101 Skeleton Dataset}

\begin{table}[t]
  \caption{\textbf{Quantitative comparison between w/o ICL and MaskICL on the Gym99-Skeleton dataset.} All values are reported in position (MPJPE, N-MPJPE) or velocity error (MPJVE).}
  \centering
  \large
  \resizebox{\columnwidth}{!}{%
    \begin{tabular}{c c c c c}
      \hline
      \textbf{Severity} & \textbf{Method} & \textbf{MPJPE$\downarrow$} & \textbf{N-MPJPE$\downarrow$} & \textbf{MPJVE$\downarrow$} \\
      \hline
      \multirow{2}{*}{\textbf{Min.}} & \cellcolor{yellow!10}w/o ICL & \cellcolor{yellow!10}0.115 & \cellcolor{yellow!10}0.103 & \cellcolor{yellow!10}0.114 \\
                                  & \cellcolor{cyan!10}\textbf{Ours} & \cellcolor{cyan!10}\textbf{0.106} & \cellcolor{cyan!10}\textbf{0.098} & \cellcolor{cyan!10}\textbf{0.047} \\
      \hline 
      \multirow{2}{*}{\textbf{Mod.}} & \cellcolor{yellow!10}w/o ICL & \cellcolor{yellow!10}0.134 & \cellcolor{yellow!10}0.115 & \cellcolor{yellow!10}0.117 \\
                                  & \cellcolor{cyan!10}\textbf{Ours} & \cellcolor{cyan!10}\textbf{0.119} & \cellcolor{cyan!10}\textbf{0.109} & \cellcolor{cyan!10}\textbf{0.085} \\
      \hline
      \multirow{2}{*}{\textbf{Sev.}} & \cellcolor{yellow!10}w/o ICL & \cellcolor{yellow!10}0.169 & \cellcolor{yellow!10}0.140 & \cellcolor{yellow!10}0.121 \\
                                  & \cellcolor{cyan!10}\textbf{Ours} & \cellcolor{cyan!10}\textbf{0.147} & \cellcolor{cyan!10}\textbf{0.132} & \cellcolor{cyan!10}\textbf{0.113} \\
      \hline
    \end{tabular}%
  }
  \label{tab:appendix_maskicl_mpjpe}
\end{table}

\begingroup
\setlength{\tabcolsep}{1.6pt}
\renewcommand{\arraystretch}{1.2}
\footnotesize
\begin{table}[t]
  \centering
  \caption{\textbf{Ablation Study on multi-GCNs and GCN types.} Three levels of temporal corruption were set: minor, moderate, and severe, performed on the Gym288-skeleton dataset. The evaluation metrics include top-1 accuracy and top-5 accuracy.}
  \begin{tabular}{lcccccc}
    \toprule
    & \multicolumn{2}{c}{\textbf{G288-Min.}} & \multicolumn{2}{c}{\textbf{G288-Mod.}} & \multicolumn{2}{c}{\textbf{G288-Sev.}} \\
    \cmidrule(lr){2-3}\cmidrule(lr){4-5}\cmidrule(lr){6-7}
    \multirow{-3}{*}{\textbf{Method}}  & \textbf{Top-1} & \textbf{Top-5} & \textbf{Top-1} & \textbf{Top-5} & \textbf{Top-1} & \textbf{Top-5} \\
    \midrule
    ST-GCN-based       & 0.778 & 0.923 & 0.756 & 0.913 & 0.723 & 0.892 \\
    Multi-GCNs   & 0.808 & 0.925 & \textbf{0.801} & 0.926 & \textbf{0.791} & \textbf{0.924} \\
    \midrule
    \rowcolor{cyan!10}
    \textbf{FineTec (Ours)} & \textbf{0.815} & \textbf{0.934} & 0.797 & \textbf{0.928} & 0.781 & 0.918 \\
    \bottomrule
  \end{tabular}
  \label{tab:appendix_multi_gcn}
\end{table}
\endgroup

To further assess the generalization and robustness of our proposed method, we conducted additional experiments on the UCF101 skeleton dataset. 
The sequence length is set to 16.
As presented in Table~\ref{tab:appendix_results_ucf101}, we compare our method, FineTec, with other baselines under three levels of temporal corruption: minor, moderate, and severe. 
FineTec achieves the highest Top-1 accuracy under both minor (0.652) and moderate (0.638) corruption levels. 
Even under severe corruption, our method maintains competitive performance, significantly outperforming ST-GCN. 
These results underscore the strong generalization ability and robustness of our method in large-scale action recognition scenarios.

\subsection{Ablation on ICL scheme in Skeletal Restoration}
As presented in Table~\ref{tab:appendix_maskicl_mpjpe}, our module demonstrates superior performance against the baseline without in-context learning (w/o ICL) on the Gym99 Skeleton dataset. 
Specifically, even under severe perturbations (Sev), ours achieves notable reductions in position and velocity errors, with MPJPE decreasing from 0.169 to 0.147, N-MPJPE from 0.140 to 0.132, and MPJVE from 0.121 to 0.113. 
These results demonstrate that our designed masking mechanism and in-context learning effectively improve joint estimation accuracy and enhance the overall  robustness of pose representation under temporal corruptions.

\subsection{Analysis of Per-Class Prediction Performance on Gym99 and Gym288}

\begin{table}[t]
\centering
\caption{\textbf{Model Complexity and Inference Speed.} Comparison of parameter counts (in millions) and inference time (in seconds) on the Gym288-skeleton dataset.}
\label{tab:appendix_model_stats}
\footnotesize
\setlength{\tabcolsep}{4pt}
\renewcommand{\arraystretch}{1.2}
\begin{tabular}{lcc}
\toprule
\textbf{Model} & \textbf{Params (M)} & \textbf{Time (s)} \\
\midrule
ST-GCN~\cite{yan2018ST-GCN} & 11.04 & 103 \\
AAGCN~\cite{shi2020skeleton} & 13.42 & 186 \\
CTRGCN~\cite{chen2021channel} & 5.31 & 220 \\
\midrule
\rowcolor{cyan!10}
\textbf{FineTec (Ours)} & 10.51 & 169 \\
\bottomrule
\end{tabular}
\end{table}

As detailed in Figure~\ref{fig:appendix_gym99_recognition_analysis} and Figure~\ref{fig:appendix_gym288_recognition_analysis}, we conducted a per-class performance analysis on the Gym99 and Gym288-skeleton datasets. 
On Gym99, our model achieves an overall accuracy of 89.99\%. 
Notably, four categories (91, 51, 8, and 65) reach 100\% accuracy, indicating robust feature extraction for these actions. 
However, some classes exhibit significantly lower performance, such as category 86 (\textit{i.e.}, Uneven Bars: Clear pike circle backward to handstand) with only 28.36\% accuracy. 
This discrepancy is likely attributable to a limited number of training samples (67 for category 86) and high inter-class similarity, as its motion patterns closely resemble those of categories 75 and 72, making differentiation challenging.

The Gym288-skeleton dataset poses a significant challenge, evidenced by a modest overall accuracy of 81.14\%. This difficulty is attributable to several factors: 1) a threefold increase in action categories to 288, which heightens inter-class similarity; 
2) a severe class imbalance, leading to degraded performance on under-represented categories; 
and 3) the intrinsic complexity of actions, characterized by intricate spatiotemporal patterns and subtle variations that demand advanced modeling capabilities. 
Despite these hurdles, our purposed FineTec demonstrates robustness by achieving perfect accuracy on several well-populated classes. 
These findings validate Gym288-skeleton as a more rigorous benchmark for evaluating model generalization, owing to its fine-grained class distinctions and heightened action similarities.

\subsection{Analysis of Multi-GCNs and GCN types}

To investigate the impact of the GCN recognition module on our framework, we first evaluated a standard ST-GCN~\cite{yan2018ST-GCN} as the recognition backbone. 
Subsequently, we experimented with a Multi-GCN architecture, where six parallel GCNs individually process each of the six sequence types (\textit{i.e.}, displacement $S_{base}, S_{dyna}, S_{stat}$ and accelerate $a_{base}, a_{dyna}, a_{stat}$), followed by a fusion of the extracted features. 

As shown in Table~\ref{tab:appendix_multi_gcn}, the Multi-GCN approach consistently outperforms the single ST-GCN across all levels of temporal corruption on the Gym288-skeleton dataset. 
Specifically, for minor, moderate, and severe corruption, the Multi-GCN architecture improves the Top-1 accuracy by 3.0\%, 4.5\%, and a substantial 6.8\%, respectively, compared to the standard ST-GCN. 
This trend highlights that dedicating separate GCN modules to different sequence types yields a more robust representation, with the performance advantage becoming more pronounced as the temporal data quality degrades. 
While our proposed FineTec method achieves the highest accuracy under minor corruption, the significant and consistent gains of the Multi-GCN architecture over the baseline ST-GCN validate its effectiveness in handling diverse and corrupted temporal sequences.

\subsection{Comparison of Models’ Parameter Counts and Inference Time}
As detailed in Table~\ref{tab:appendix_model_stats}, our proposed model, FineTec, demonstrates a compelling balance between computational complexity and inference efficiency. 
With 10.51 million parameters, FineTec is more lightweight and faster than AAGCN, which has 13.42 million parameters and an inference time of 186 seconds. 
Notably, while CTRGCN features a smaller parameter count at 5.31 million, our model achieves a significantly faster inference time of 169 seconds compared to CTRGCN's 220 seconds. 
This highlights FineTec's effective design in achieving a favorable trade-off, ensuring competitive performance without excessive computational overhead.

\subsection{Limitation and Future Works}

While FineTec demonstrates promising results in fine-grained temporal defect detection by focusing on the explicit modeling of skeletal spatial and temporal dynamics, we acknowledge this as a preliminary step. There are several limitations and avenues for future research. 
1) The primary innovation of this work lies in the explicit parsing of skeleton sequences, rather than a fundamental redesign of the GCN topology . 
A compelling future direction would be to embed the physical laws and biomechanical principles emphasized in this paper directly into the topological structure of GCNs. 
2) Although this paper introduces a novel task, fine-
grained temporal-corrupted action recognition, its scope is currently limited to the single task of action recognition. 
In our future work, we plan to leverage the proposed fine-grained explicit parsing methodology and apply it to a broader spectrum of skeleton-based tasks, such as skeleton completion and future frame prediction.

\end{document}